\documentclass{article}

    \usepackage[nonatbib,final]{mlsafety_neurips_2022}

\usepackage{graphicx}
\usepackage{algorithm}
\usepackage{subcaption}
\usepackage{float}
\usepackage{amsmath, amssymb}
\usepackage{mathtools}
\usepackage{appendix}
\usepackage{todonotes}
\usepackage{soul}
\usepackage{fixltx2e}
\usepackage{multirow}

\usepackage[utf8]{inputenc} 
\usepackage[T1]{fontenc}    
\usepackage{hyperref}       
\usepackage{url}            
\usepackage{booktabs}       
\usepackage{amsfonts}       
\usepackage{nicefrac}       
\usepackage{microtype}      
\usepackage{xcolor}         

\usepackage{bbm}
\usepackage[font=small,labelfont=bf]{caption}

\title{\centering{Boundary Adversarial Examples\\Against Adversarial Overfitting}}

\author{%
  Muhammad Zaid Hameed \\
  IBM Research Europe\\
  Dublin, Ireland \\
  \texttt{Zaid.Hameed@ibm.com} \\
   \And
   {Beat Buesser} \\
  IBM Research Europe\\
  Dublin, Ireland \\
   \texttt{beat.buesser@ie.ibm.com} \\
}

\newcommand{\R}{\mathbb{R}}
\DeclareMathOperator*{\argmax}{arg\,max}

\newcommand{\Y}{\mathcal{Y}}
\newcommand{\btheta}{{\boldsymbol\theta}}

\newcommand{\bx}{\mathbf{x}}
\newcommand{\PGD}{\mathrm{PGD}}
\newcommand{\PGDTen}{\mathrm{{PGD}_{10}}}
\newcommand{\PGDTwe}{\mathrm{{PGD}_{20}}}

\newcommand{\ATBS}{AT-TE\textsubscript{BS}}
\newcommand{\MLCATBS}{MLCAT-WP+TE\textsubscript{BS}}
\newcommand{\ATCS}{AT-TE\textsubscript{CS}}
\newcommand{\MLCATCS}{MLCAT-WP+TE\textsubscript{CS}}
\newcommand{\MLCATXB}{MLCAT-WP+XE\textsubscript{BS}}

\begin{document}

\maketitle

\begin{abstract}
Standard adversarial training approaches suffer from robust overfitting where the robust accuracy decreases when models are adversarially trained for too long. The origin of this problem is still unclear and conflicting explanations have been reported, i.e., memorization effects induced by large loss data or because of small loss data and growing differences in loss distribution of training samples as the adversarial training progresses. Consequently, several mitigation approaches including early stopping, temporal ensembling and weight perturbations on small loss data have been proposed to mitigate the effect of robust overfitting. However, a side effect of these strategies is a larger reduction in clean accuracy compared to standard adversarial training. In this paper, we investigate if these mitigation approaches are complimentary to each other in improving adversarial training performance. We further propose the use of \textit{helper adversarial examples} that can be obtained with minimal cost in the adversarial example generation, and show how they increase the clean accuracy in the existing approaches without compromising the robust accuracy.
\end{abstract}

\section{Introduction}

Adversarial examples have gained great attention in ML research \cite{szegedy2013intriguing,goodfellow2014explaining,carlini2017towards} and building models with increased robustness against such examples is still an open challenge \cite{madry2017towards,zhang2019theoretically,shafahi2019adversarial,qin2019local,hydra2020sehwag,wu2020adversarial,gowal2020uncovering, chen2021robust}. Adversarial training (AT) \cite{madry2017towards} is the most successful approach against adversarial examples where approximate worst-case adversarial examples are used to train the model. \\
The success of AT is impeded by high computational costs compared to standard training and gaps between robust and clean accuracy \cite{madry2017towards,zhang2019theoretically,tsipras2018robustness,yang2020closer}. It has been shown that AT reduces the clean accuracy \cite{zhang2019theoretically} and mitigation techniques based on label smoothing and stochastic weight averaging \cite{chen2021robust} or use of additional unlabeled data \cite{unlabeled_data_2019} have been proposed, which solve the problem only partially. \\
Recently, another phenomenon associated with adversarial training has been discovered and coined \textit{robust overfitting} \cite{rice2020overfitting,chen2021robust,dong2022exploring,yu2022understanding}, which occurs when models are adversarially trained for too long. Especially after the first learning rate decay, the robust accuracy tends to decrease with additional training. Even though overfitting to training data results in increased clean accuracy, the robust accuracy is negatively impacted by this additional training. The exact causes of robust overfitting are still unclear and most explanations are conflicting. For example, memorization effects induced by large loss data, which give rise to high confidence predictions, have been shown to be a driving force behind robust overfitting \cite{dong2022exploring}. On the other hand, small loss data and large differences in loss distribution across training samples, tend to amplify as AT progresses, and have been shown to enhance robust overfitting \cite{yu2022understanding}. Reported mitigation approaches like temporal ensembling (TE) \cite{dong2022exploring} and model weight perturbation (WP) \cite{yu2022understanding} alleviate robust overfitting, but tend to trade off more clean accuracy when compared to standard adversarial training (AT) \cite{madry2017towards}. This shows that increased regularization induced by these schemes negatively impact the generalization performance on clean data.\\ 
Here, we analyze how these two seemingly conflicting approaches, TE and WP, which attribute robust overfitting to training with samples with large and small loss data, respectively, mitigate robust overfitting and if they can be combined to create a compounding mitigation effect. First, we investigate how TE and WP behave during AT and find that they induce different types of regularization on adversarial examples during overfitting. Second, we evaluate a combination of TE and WP and find that there is no compounding improvement in robust accuracy. Third, we minimize loss on clean input data along-with these robust overfitting mitigation approaches which results in increased clean accuracy at the expense of robust accuracy. Thus, we find that the negative impacts on clean accuracy induced by TE and WP can be prevented without affecting the robust accuracy with appropriate regularization.\\ 
We propose to employ TE/WP based robust overfitting mitigation with the additional use of helper adversarial examples along-with a TE-based regularization term and demonstrate improvement in clean accuracy over the existing AT approaches. The helper examples can be created during the adversarial example generation in AT with a minimal computational overhead of just one extra forward pass. Our idea is to use perturbed samples close to decision boundaries as helper examples with their property of increasing clean accuracy with only a small degradation in robust accuracy compared to using clean data only.

\vspace{-3mm}

\section{Preliminaries}
\vspace{-2mm}
Adversarial training (AT) has empirically proved to be the most effective defense strategy against adversarial evasion attacks \cite{madry2017towards, zhang2019theoretically,shafahi2019adversarial,gowal2020uncovering,chen2021robust}. In the following we will briefly describe the mechanics of AT \cite{madry2017towards} and two robust mitigation schemes \cite{dong2022exploring, yu2022understanding}.
\vspace{-2mm}
\subsection{Adversarial Training (AT) \cite{madry2017towards}}
\vspace{-2mm}
We consider a classification problem where an input sample $\bx \in \R^n$ belongs to a true class $y$ among a set of classes $\Y =\{1, 2, \ldots, Y\}$. A DNN-based classifier $F_\btheta: \R^n \times \Y \to \R$, assigns a label  $\hat{y} \in \argmax_{y \in \Y} F_\btheta(\bx,y)$ to $\bx$, where $\btheta \in \Theta$ denotes the parameters of DNN. With a slight abuse of notation, we also use $F$ to denote the classifier, $F(\bx)$ to denote the class label assigned to $\bx$, and $F(\bx,y)$ to denote the score of class $y$ for input $\bx$ and $p(\bx)$ denotes the DNN prediction probability vector. For a correctly classified input $\bx$ i.e., $y = F(\bx)$ is the true label, an adversarial attack aims to find a sample $\tilde{\mathbf{x}}$ in the $\epsilon$-neighborhood of $\bx$ in norm $p$ denoted by $\mathcal{B}_{\epsilon}(\mathbf{x})=\left\{ \tilde{\mathbf{x}}: \| \tilde{\mathbf{x}}-\mathbf{x}\|_{p}  \leq \epsilon \right\}$, such that $F(\tilde{\bx}) \neq F(\bx)$. 
In practice, these adversarial examples are generated by modifying the input $\bx$ through optimizing a loss function $\mathcal{L}$ on the classifier  \cite{carlini2017towards,moosavi2016deepfool,chen2017ead,laidlaw2019functional,madry2017towards}. Finally, AT can be considered as a robust  optimization problem:
\begin{equation}
\min_{\boldsymbol{\theta}} \max_{\tilde{\mathbf{x}}\in \mathcal{B}_{\epsilon}(\bx)} \mathcal{L}(F(\tilde{\mathbf{x}},y),y),
\end{equation}
where $\tilde{\mathbf{x}}$ is the adversarial example. Hence, during training, adversarial attacks first generate adversarial examples in the neighborhood $\mathcal{B}_{\epsilon}(\mathbf{x})$ of input $\bx$ to approximately maximize the loss $\mathcal{L}$ (e.g., cross-entropy loss), followed by training on the examples to update the network parameters and achieving robustness against adversarial examples.
\vspace{-3mm}
\subsection{Adversarial Training with Temporal Ensembling (AT-TE) \cite{dong2022exploring}}
\vspace{-2mm}
Robust overfitting is attributed to neural networks training on high loss input data and over-confident predictions (memorization) by the model 
during AT \cite{chen2021robust,dong2022exploring}. To prevent this, a regularization based on temporal ensembling (TE) has been proposed in \cite{dong2022exploring} which penalizes over-confident predictions. Let $z(\bx)$ denote the ensemble prediction by a model on input $\bx$ which is updated as $z(\bx) = \eta \cdot z(\bx) + (1- \eta) \cdot p(\bx)$ in each epoch and the AT objective becomes 
\begin{equation}
\min_{\boldsymbol{\theta}} \max_{\tilde{\mathbf{x}}\in \mathcal{B}_{\epsilon}(\bx)} \{\mathcal{L}(F(\tilde{\mathbf{x}},y),y) + w \cdot \| p(\tilde{\bx}) - \hat{z}(\bx) \|_{2}^{2}\},
\end{equation}
where $\hat{z}(\bx)$ is obtained by normalizing $z(\bx)$ and $w$ is regularization weight. In this AT, 
 
the regularization term is activated close to the first learning rate decay, and prevents the network from assigning high confidence to samples with large loss.  

\vspace{-3mm}
\subsection{Minimum Loss Constrained Adversarial Training (MLCAT-WP) \cite{yu2022understanding}}
\vspace{-2mm}

Training samples with small loss and large differences in loss distribution among samples are shown to cause robust overfitting in \cite{yu2022understanding} and adversarial weight perturbation \cite{wu2020adversarial} is employed to increase the loss on such samples, which helps to mitigate robust overfitting. The training objective becomes 
\begin{equation}
\min_{\boldsymbol{\theta}}
\max_{\mathbf{v} \in \mathcal{V}}
\max_{\tilde{\mathbf{x}}\in \mathcal{B}_{\epsilon}(\bx)} \mathcal{L}(F_{\theta + v}(\tilde{\mathbf{x}},y),y),
\end{equation}
where $v \in \mathcal{V}$ is an adversarial weight perturbation and is generated by  
$\max_{\mathbf{v} \in \mathcal{V}} \sum_{i} \mathbbm{1}_{\{\mathcal{L}_{i} \leq L_{min}\}} \cdot \mathcal{L}_{i} ,
$ where $\mathcal{L}_{i} =\mathcal{L}(F_{\theta + v}(\tilde{\mathbf{x}_{i}},y_{i}),y_{i})$, $L_{min}$ is threshold for minimum adversarial loss and $\mathbbm{1}_{c}$ is an indicator function which is 1 only when condition $c$ is true.

\vspace{-3mm}
\section{Comparison of AT, AT-TE and MLCAT-WP}
\vspace{-2mm}

\begin{figure}[h!]
\vspace{-3mm}
\subfloat[Test Data Accuracy]{\includegraphics[width=3.5cm]{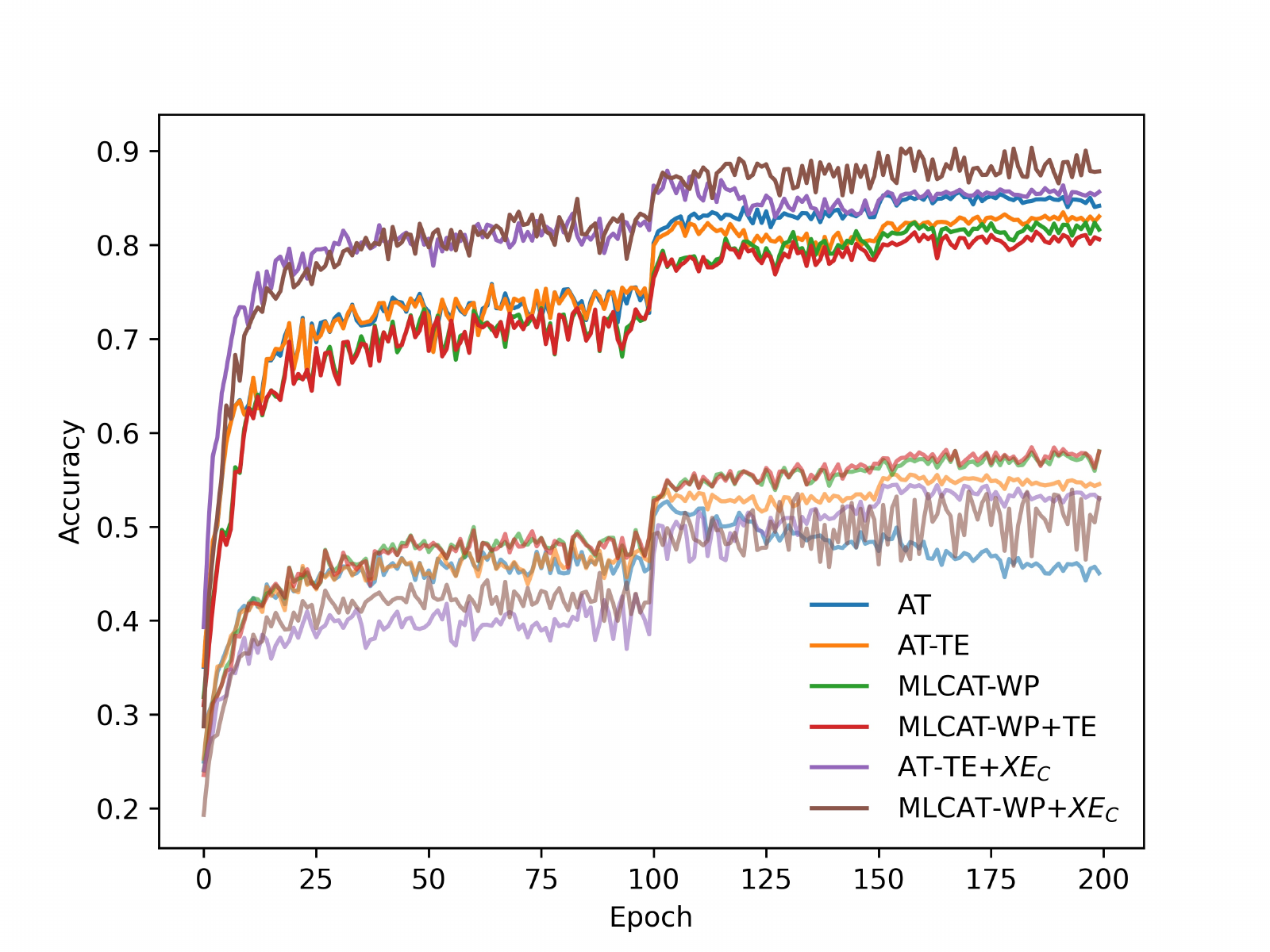}}
\subfloat[TCP]{\includegraphics[width=3.5cm]{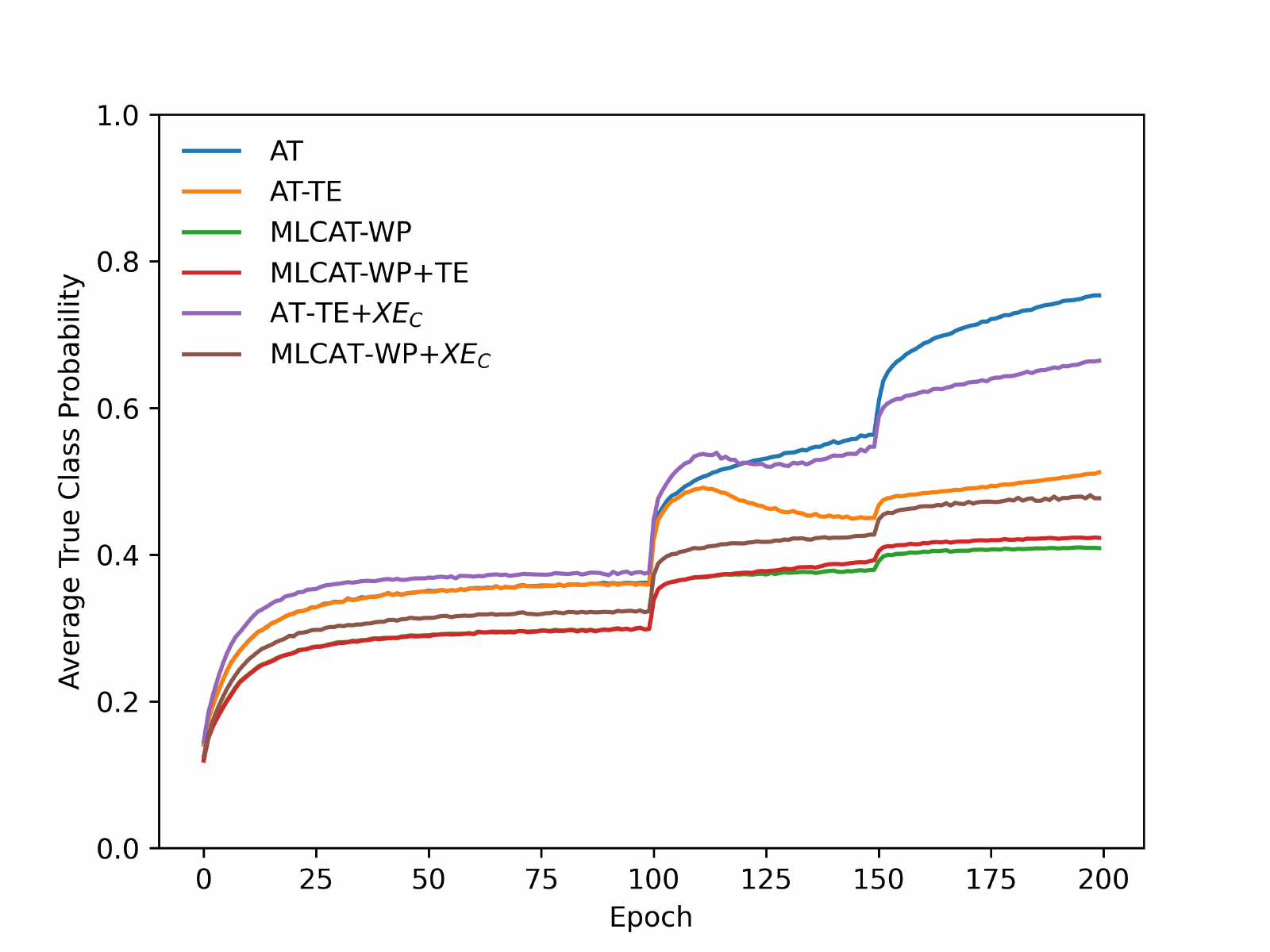}}
\subfloat[TCP, loss $\in$ [0, 0.5)]{\includegraphics[width=3.5cm]{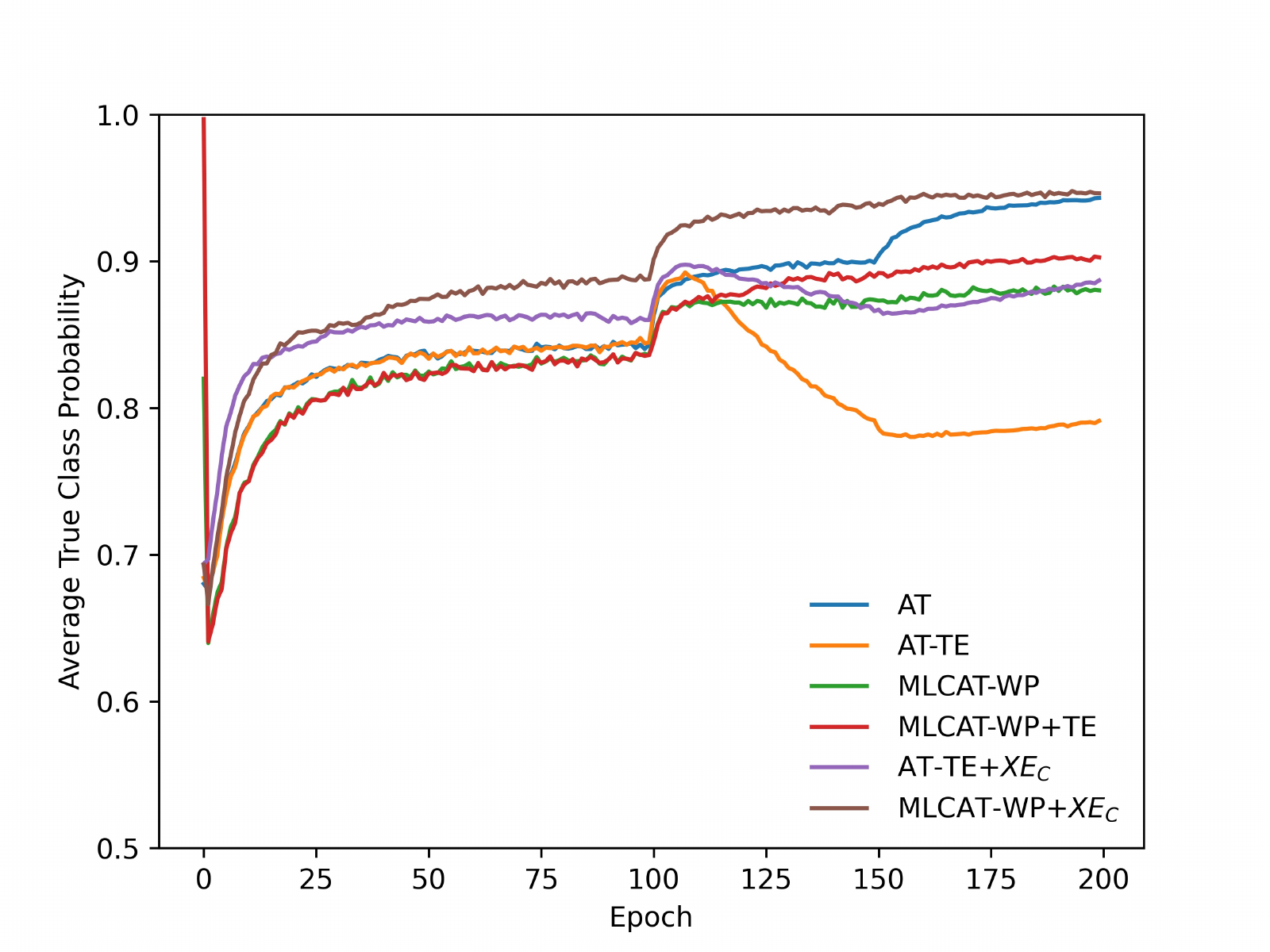}}
\subfloat[Sample Proportion, loss $\in$ [0, 0.5)]{\includegraphics[width=3.5cm]{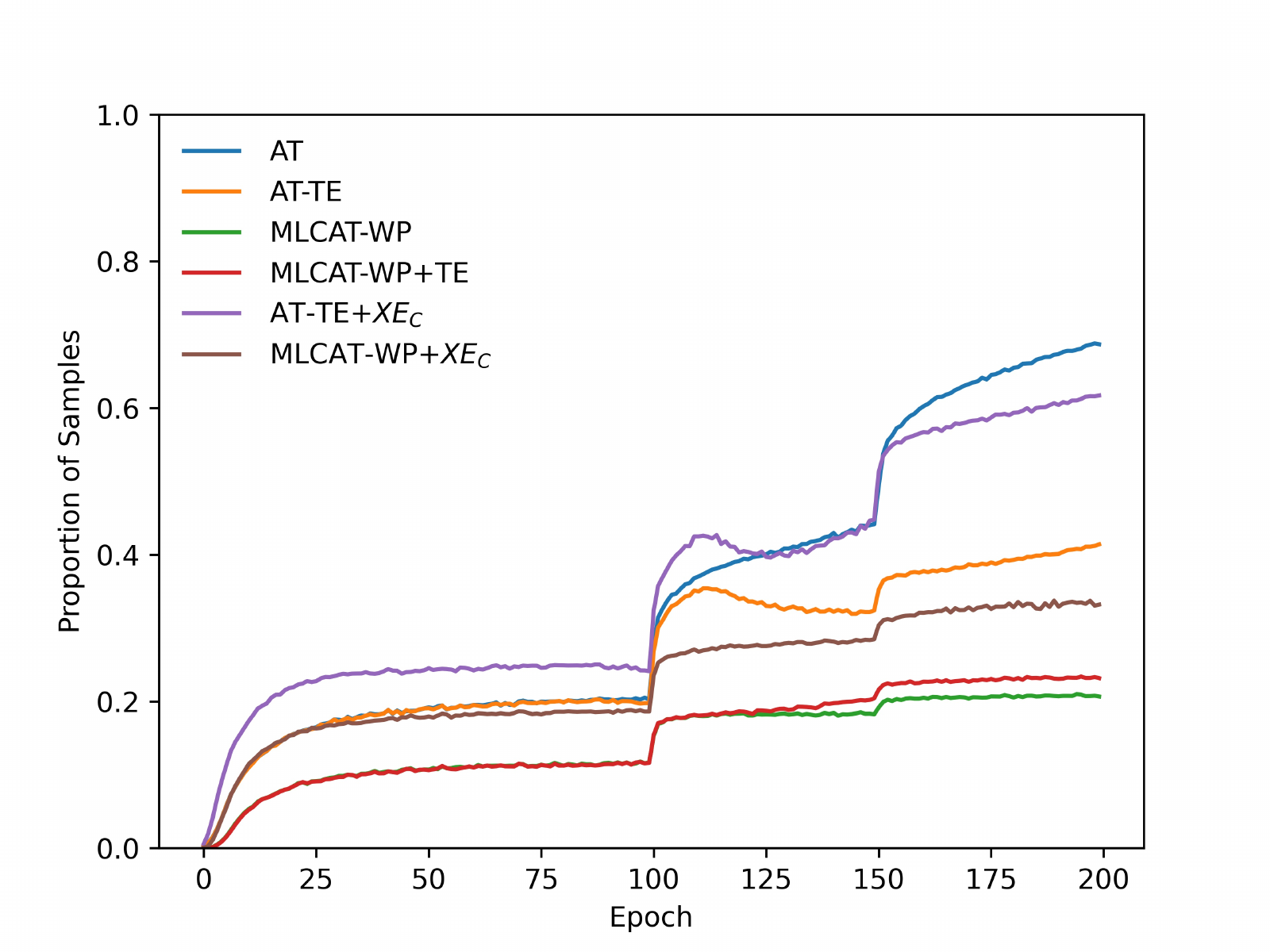}}\vspace{-1mm}

\caption{CIFAR-10 training for ResNet-18. (a) Test accuracy against clean data (dark solid lines) and $\PGDTwe$ attack (dim solid lines) are plotted.}
\label{fig1s}
\vspace{-6mm}
\end{figure}

In order to investigate how standard AT and robust overfitting mitigation approaches AT-TE and MLCAT-WP behave as training progresses, we consider image classification on CIFAR-10 \cite{CIFAR10} with a ResNet-18 model \cite{he2016deep}. 
For all approaches, we use a Projected Gradient Descent ($\PGD$) attack \cite{madry2017towards} for 10 steps denoted by $\PGDTen$ in training and $\PGDTwe$ for evaluation; see Appendix~\ref{more-exp-setup} for full details.
We observe from Figure~\ref{fig1s}(b)-(d) that as the training progresses, an AT model starts to classify training data with high true class probability (TCP) (Average TCP $\geq 0.5$) after the first learning rate decay at epoch 100 and proportion of training samples with very small loss value (in range [0, 0.5)) and high TCP also grows ($\geq 40\%$ of all training samples). On the other hand, an MLCAT-WP trained model's average TCP on training data increases gradually and stays small ($\approx 0.4$) at epoch 200 and the proportion of samples in the loss range [0, 0.5) amounts to only 20\% of all training data which shows weight perturbation prevents the model from assigning very high TCP to training data and fitting samples with small loss. In case of AT-TE the average TCP for samples with loss in range [0, 0.5) starts to decrease when temporal ensembling is activated (epoch $\geq 90$) and even though the proportion of samples increase, it lies between AT and MLCAT-WP.  
Finally, we can also make this observation from Figure~\ref{fig1s}(a) that combining weight perturbation and temporal ensembling (MLCAT-WP+TE) does not result in any improvement in robust accuracy and its behavior closely resembles MLCAT-WP. Moreover, modifying AT-TE and MLCAT-WP to AT-TE+XE\textsubscript{C} and MLCAT-WP+XE\textsubscript{C} by including cross entropy loss on clean data results in higher clean accuracy at the cost of reduced robust accuracy and results in increase in TCP and proportions of samples with small loss.

\section{Boundary Adversarial Examples for Improving Adversarial Training}
\vspace{-3mm}
To counter the negative effect of regularization in AT-TE and MLCAT-WP on clean accuracy and negative effect of using clean sample on robust accuracy, we propose to extract additional useful information from the adversarial examples generation process in adversarial training. More specifically, we extract intermediate adversarial examples that are close to a decision boundary as soon as the perturbed sample is misclassified. Our underlying idea is that using boundary (intermediate and weak) adversarial examples in place of clean samples will guide the network to attain better clean accuracy without affecting robust accuracy too much. Thus, using this intermediate perturbed sample $\mathbf{x}'\in \mathcal{B}_{\epsilon}(\bx)$ and a regularization based on its prediction $p(\bx')$ and ensemble prediction $z(\bx')$, the AT-TE objective becomes         
\begin{equation}
\label{eq: bste}
\min_{\boldsymbol{\theta}} \{ \mathcal{L}(F(\mathbf{x}',y),y) + w \cdot \| p(\bx') - \hat{z}(\bx') \|_{2}^{2}\ + \max_{\tilde{\mathbf{x}}\in \mathcal{B}_{\epsilon}(\bx)} \{\mathcal{L}(F(\tilde{\mathbf{x}},y),y) + w \cdot \| p(\tilde{\bx}) - \hat{z}(\bx) \|_{2}^{2}\}\}  ,
\end{equation}
where $\hat{z}(\bx')$ is  normalized $z(\bx')$, and $z(\bx')$ is updated as $z(\bx') = \eta \cdot z(\bx') + (1 - \eta ) \cdot p(\bx')$ in each epoch. We denote this modified AT-TE as \ATBS. Similarly, MLCAT-WP objective becomes \MLCATBS \ by including TE and boundary sample as 
\begin{equation}
\label{eq: bswp}
\min_{\boldsymbol{\theta}} \{  \mathcal{L}(F_{\theta + v}(\mathbf{x}',y),y) + w \cdot \| p(\bx') - \hat{z}(\bx') \|_{2}^{2} \ + \ \max_{\mathbf{v} \in \mathcal{V}} \max_{\tilde{\mathbf{x}}\in \mathcal{B}_{\epsilon}(\bx)} \{\mathcal{L}(F_{\theta + v}(\tilde{\mathbf{x}},y),y) + w \cdot \| p(\tilde{\bx}) - \hat{z}(\bx) \|_{2}^{2}\}\} 
\end{equation}

\vspace{-3mm}
\section{Experimental Evaluation}\label{ss:nr}
\vspace{-2mm}

\begin{figure}[h!]
\center
\subfloat[CIFAR-10 Accuracy]{\includegraphics[width=4.8cm]{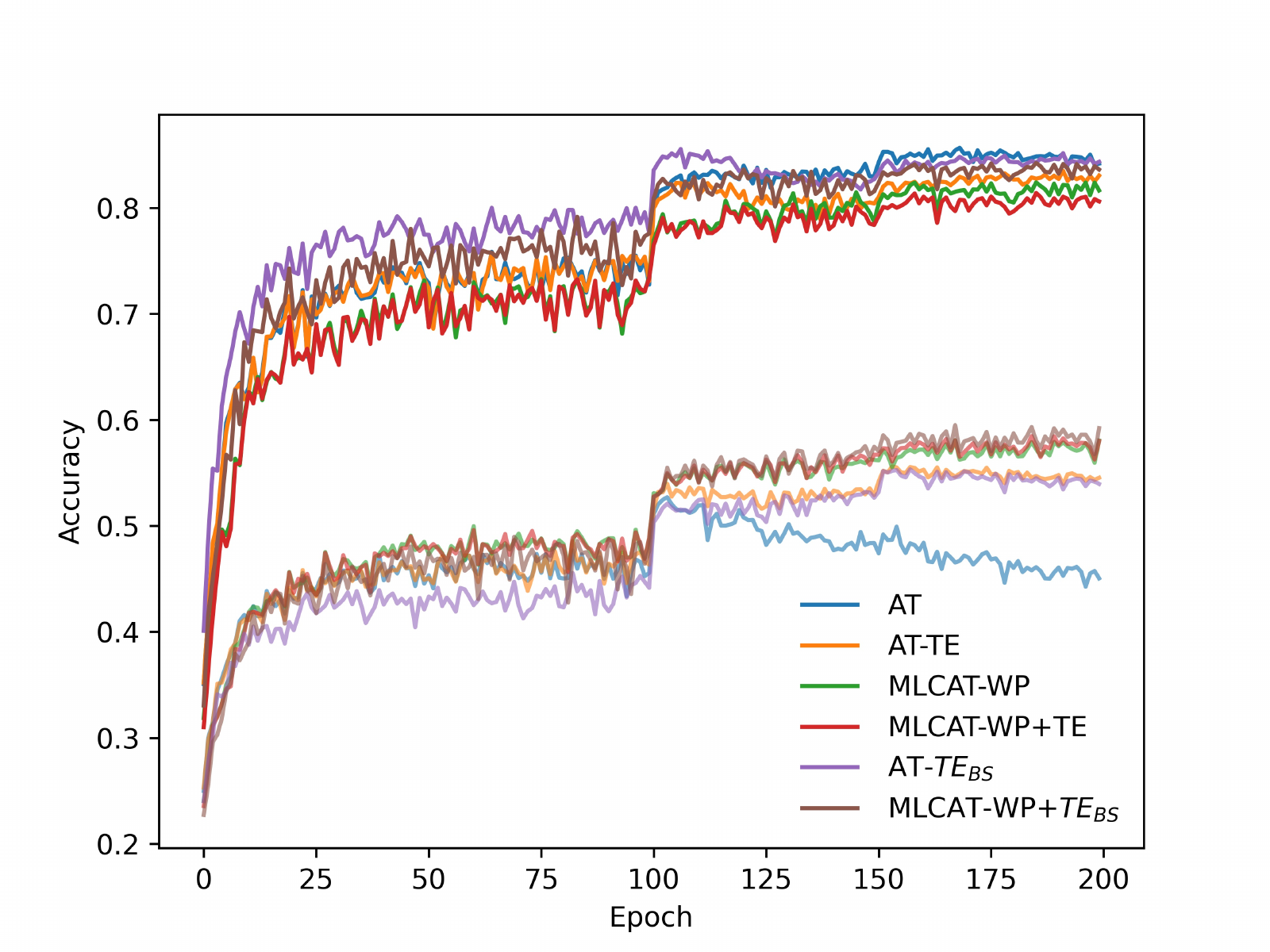}}
\subfloat[CIFAR-100]{\includegraphics[width=4.8cm]{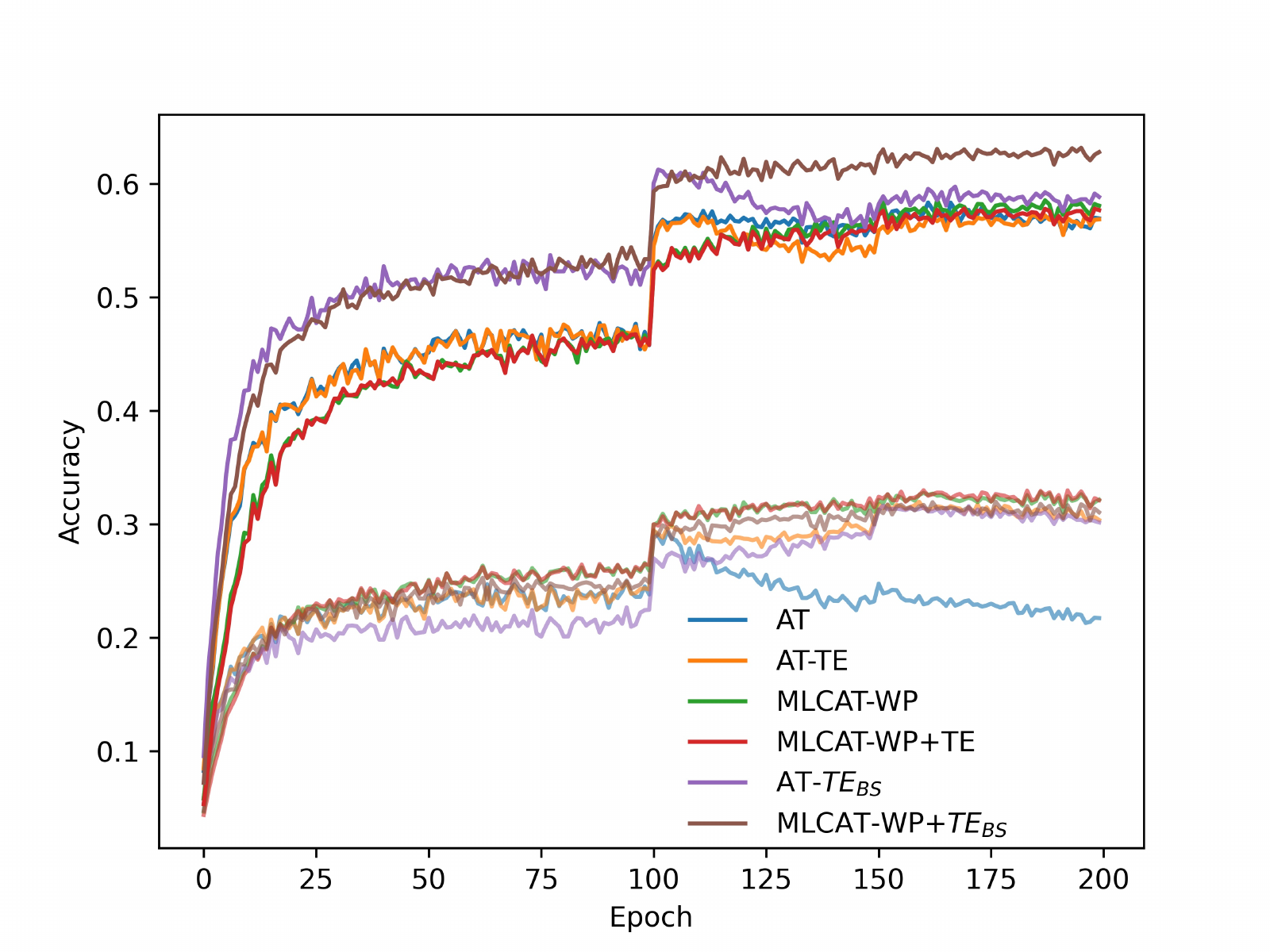}}
\subfloat[SVHN]{\includegraphics[width=4.8cm]{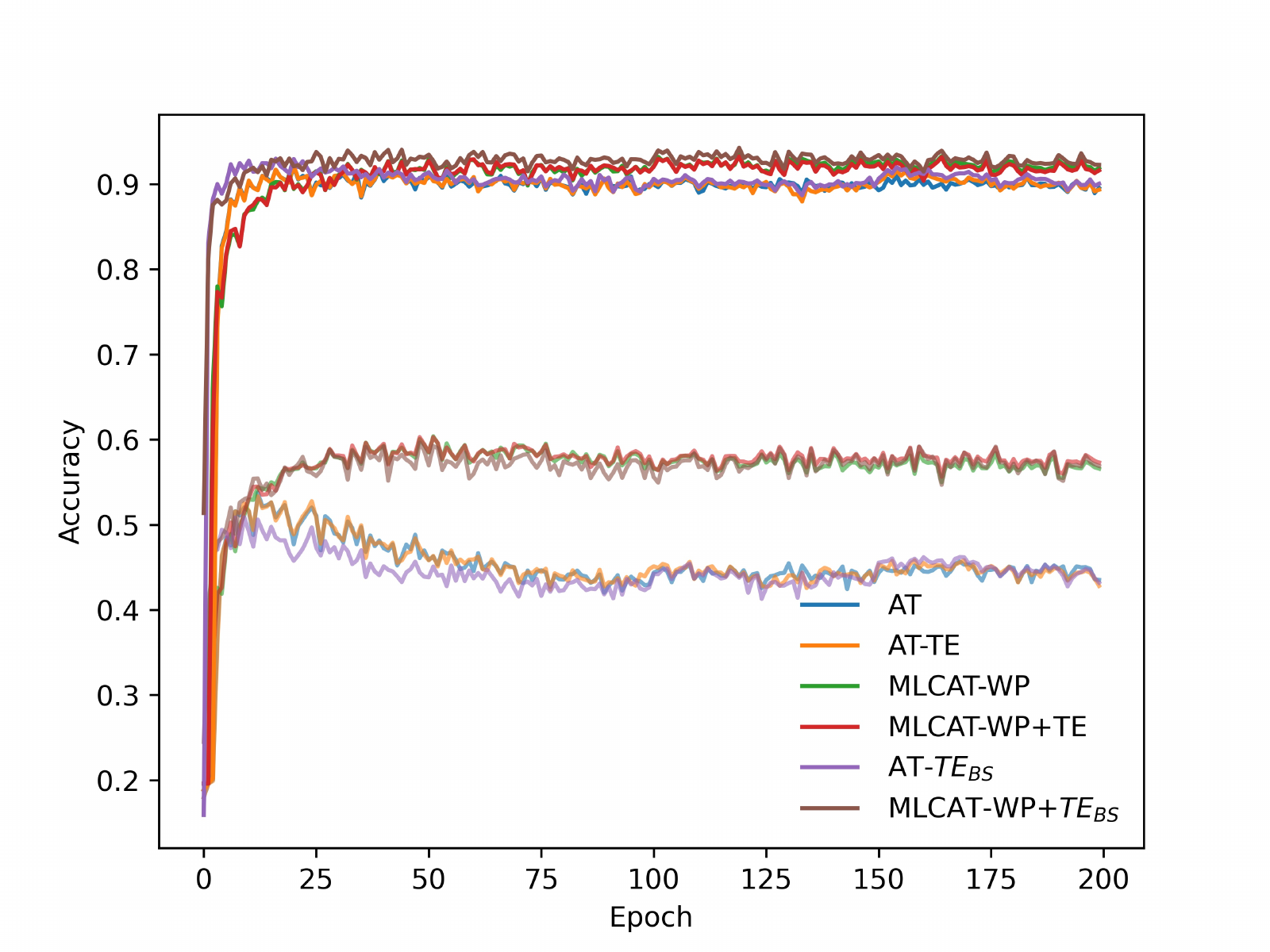}}
\vspace{-1mm}
\caption{Accuracy results for AT using ResNet-18. Clean test accuracy (dark solid lines) and $\PGDTwe$ attack test accuracy (dim solid lines) are plotted.}
\label{ee:fig1}
\vspace{-6mm}
\end{figure}

We train a ResNet-18 model using AT, AT-TE, MLCAT-WP, MLCAT-WP+TE, \ATBS \ and \MLCATBS \ for CIFAR-10 \cite{CIFAR10}, CIFAR-100 \cite{CIFAR10} and SVHN \cite{netzer2011reading}; see Appendix~\ref{more-exp-setup} for details. We use $\PGDTen$ attack ($\epsilon=8/255$, $L_{\infty}$ norm)  during training and $\PGDTwe$ at inference. In addition, we also run AutoAttack (AA) \cite{croce2020reliable} which is an ensemble of different attacks for a more reliable evaluation. Table~\ref{Tab: ee1} shows that both \ATBS \ and \MLCATBS \ attain significant increase in clean accuracy over their counterparts AT-TE and MLCAT-WP/MLCAT-WP+TE ($\approx$2\%-3\% in CIFAR-10, 2\%-5\% in CIFAR-100, and 1\%-2\% in SVHN). AT-TE and \ATBS \ do not prevent overfitting in the SVHN dataset because the training data fits very early in training with high confidence, whereas temporal ensembling activates closer to first learning rate decay and thus, regularization based on ensemble prediction is ineffective. On the other hand, weight perturbation based approaches MLCAT-WP and \MLCATBS \ result in superior performance across all datasets compared to AT-TE and \ATBS \ in terms of clean and robust accuracy; and use of adversarial boundary examples significantly boosts the clean accuracy especially for CIFAR-100 and SVHN datasets in \MLCATBS.  
From Figure~\ref{ee:fig1}, we observe that for all three datasets, \ATBS \ and \MLCATBS approximately retain the robust accuracy of AT-TE and MLCAT-WP, but increase the clean accuracy to match and even surpass AT in case of the CIFAR-100 and SVHN datasets.

\begin{table}
  \caption{Test accuracy (mean and standard deviation of 5 runs). ``Last" and ``Best" refer to test accuracy at the end of training, and end of epoch that gives the highest accuracy w.r.t. test data respectively.}
  \label{Tab: ee1}
  \centering
\resizebox{0.8\columnwidth}{!}{%
  \begin{tabular}{llllllll}
    \toprule
    Dataset &   Name    &   Clean(Best) &   Clean(Last)    &  Robust(Best)  &   Robust(Last)    &   AA(Best)    &   AA(Last) \\
    \midrule
    \multirow{6}*{CIFAR-10} &AT & 82.03$\pm0.42$  & 84.37$\pm$0.30 & 52.64$\pm$0.13& 45.06$\pm$0.70 &    48.04$\pm$0.15       &  42.64$\pm$0.62    \\
    & AT-TE     & 82.11$\pm$0.14  &    82.73$\pm$0.22   & 55.69$\pm$0.11  &   54.15$\pm$0.42    &  49.99$\pm$0.16   &  48.95$\pm$0.37 \\
    & MLCAT-WP  & 82.05$\pm$0.31   & 81.78 $\pm$0.26 & 58.16$\pm$0.13 & 57.46$\pm$0.45  &   50.43$\pm$0.07   &  49.96$\pm$0.31    \\
    & MLCAT-WP+TE   & 81.06$\pm$0.49  &   80.74$\pm$0.24    &   58.48$\pm$0.08    &   57.71$\pm$0.22    & \textbf{50.67$\pm$0.10}   &  \textbf{50.46$\pm$0.20}   \\
    & \ATBS     &   \textbf{83.99$\pm$0.14} & \textbf{84.46$\pm$0.36} &   55.26$\pm$0.07    &   53.58$\pm$0.74    &  50.17$\pm$0.09   &  48.80$\pm$0.70 \\
& \MLCATBS  & 83.712$\pm$0.51 &   83.75$\pm$0.41    &   \textbf{59.20$\pm$0.21}    &   \textbf{58.53$\pm$0.52}    &  50.38$\pm$0.07   &  50.30$\pm$0.26   \\
    \midrule    
    \multirow{6}*{CIFAR-100} &AT &  55.49$\pm$0.60  &   56.93$\pm$0.15  &   29.50$\pm$0.22  &   22.11$\pm$0.25 &   25.05$\pm$0.44   &   19.97$\pm$0.19     \\
    & AT-TE     &   56.43$\pm$0.25  &   56.94$\pm$0.44  &   32.07$\pm$0.06  &   30.67$\pm$0.18  &  26.19$\pm$0.18  &   25.1$\pm$0.09      \\
    & MLCAT-WP     &    57.65$\pm$0.41  &   57.96$\pm$0.30  &   32.73$\pm$0.11  &   31.68$\pm$0.61  &   \textbf{27.14$\pm$0.14}  &   26.48$\pm$0.41  \\
    & MLCAT-WP+TE     & 56.60$\pm$0.56  &   57.44$\pm$0.33  &   \textbf{33.02$\pm$0.16}  &   \textbf{32.02$\pm$0.56}  &   27.08$\pm$0.16  &   \textbf{26.58$\pm$0.54}  \\
    & \ATBS     &   58.78$\pm$0.23  &   58.82$\pm$0.29  &   31.59$\pm$0.09  &   30.40$\pm$0.16  &  25.99$\pm$0.18  &   24.98$\pm$0.23      \\
    & \MLCATBS     &    \textbf{62.50$\pm$0.28}  &   \textbf{62.45$\pm$0.42}  &   31.95$\pm$0.06  &   30.72$\pm$0.65  &   26.57$\pm${0.14} &  25.74$\pm$0.48  \\
    \midrule    
    \multirow{6}*{SVHN} &AT & 89.08$\pm$0.47  & 89.91$\pm$0.30  &   53.21$\pm$0.32  &   44.70$\pm$0.60  & 45.59$\pm$0.61    &   39.97$\pm$0.63    \\
    & AT-TE     & 89.17$\pm$0.58    &   89.83$\pm$0.34  &   53.21$\pm$0.18  &   44.24$\pm$0.71  &  45.81$\pm$0.43  &   39.65$\pm$0.81      \\
    & MLCAT-WP     &    91.45$\pm$0.26  &    91.91$\pm$0.21 &   60.25$\pm$0.28  &   56.98$\pm$0.62  &   \textbf{51.60$\pm$0.37}  &   49.20$\pm$0.30  \\
    & MLCAT-WP+TE     & 91.53$\pm$0.46  &   91.67$\pm$0.30  &   \textbf{60.33$\pm$0.30}  & \textbf{57.71$\pm$ 0.84}    &   51.57$\pm$0.31  &   \textbf{49.89$\pm$0.54}   \\
    & \ATBS     & 91.58$\pm$0.32  &   90.43$\pm$0.46  &   50.94$\pm$0.16  &   44.43$\pm$0.66    &  45.37$\pm$0.29  &   39.43$\pm$0.70      \\
    & \MLCATBS     &    \textbf{92.43$\pm$0.46}  &   \textbf{92.53$\pm$0.35}  &   60.07$\pm$0.31  &   57.61$\pm$ 0.94    &   51.27$\pm$0.49  &   49.42$\pm$0.54  \\
    \bottomrule
  \end{tabular}}
\vspace{-3mm}
\end{table}

\vspace{-3mm}
\section{Conclusion}
\vspace{-2mm}
We investigate temporal ensembling and weight perturbation for mitigating robust overfitting and discover that temporal ensembling mainly influences high confidence predictions whereas weight perturbation affects both confidence in predictions and small loss data samples. 
Overall, adversarial weight perturbation, which directly prevents the model from fitting low loss data samples, achieves better clean and robust accuracy compared to temporal ensembling. Furthermore, we propose to use samples close to decision boundary for improving clean accuracy. These can be directly obtained from the adversarial examples generation process during adversarial training with minimal additional cost. Together with ensemble prediction regularization, this helps in retaining the robust accuracy of both robust overfitting mitigation approaches but significantly increases the clean accuracy.

\bibliography{neurips_2022}
\bibliographystyle{ieeetr}


\newpage

\appendix

\section{Experimental Setup} \label{more-exp-setup}

For all experiments, we use ResNet-18 models \cite{he2016deep} which are trained using SGD with momentum value of 0.9 and weight decay of 5 × $10^{-4}$. Initial learning rate is set to 0.1 for CIFAR-10 and CIFAR-100 datasets and 0.01 for SVHN dataset, which is divided by 10 at the $100^{th}$ and $150^{th}$ epochs with a total training of 200 epochs. Data augmentation consisting of horizontal flip and random crop is used for CIFAR-10 and CIFAR-100 datasets, while no data augmentation is used for SVHN datatset. 

For $\PGD$ attack parameters for training, we set the $\epsilon = \frac{8}{255}$ in $L_{\infty}$ norm (maximum perturbation) and 10 attack steps for all datasets where step size of $\frac{2}{255}$ is used for CIFAR-10 and CIFAR-100 datatsets and step size of $\frac{1}{255}$ for SVHN.  For evaluation we consider $\PGD$ attack with 20 steps with $\epsilon = \frac{8}{255}$ in $L_{\infty}$ norm and step size of $\frac{2}{255}$ for all datasets.

For temporal ensembling based approaches, the value of temporal ensembling weight parameter $w$ is adjusted experimentally and is set to 300 in AT-TE, \ATBS \ and \ATCS \ for CIFAR-10 and SVHN datasets and 3000 for CIFAR-100 datatset along a Gaussian ramp-up curve \cite{dong2022exploring}. Similarly, for MLCAT-WP+TE, \MLCATBS \ and \MLCATCS \ $w=30$ is used in training with CIFAR-10 and SVHN datasets and \ $w=300$ is used for CIFAR-100 dataset along a Gaussian ramp-up curve. The momentum term $\eta$ in ensemble prediction update is set to 0.9 and temporal ensembling activates at $90^{th}$ epoch \cite{dong2022exploring} for all experiments involving temporal ensembling. For all experiments with MLCAT-WP, MLCAT-WP+TE, \MLCATBS \ and \MLCATCS, \ $L_{min}=1.5$ for CIFAR-10 and SVHN datasets and $L_{min}=4.0$ for CIFAR-100 dataset and other parameters are set as per the original work \cite{yu2022understanding}. 

Furthermore all experiments are run on a single NVIDIA A100 Tensor Core GPU using PyTorch version 1.11.0 on Red Hat Enterprise Linux release 8.5 operating system.   

\section{Additional Results for CIFAR-10, CIFAR-100 and SVHN Datasets} \label{app: add-res}
This section contains additional results for CIFAR-10, CIFAR-100 and SVHN datasets. We consider the case when instead of boundary sample $\bx'$ we use clean input sample $\bx$ and a regularization on network's current prediction on this clean sample $p(\bx)$ and ensemble prediction $z(\bx)$. Thus \ATBS \ modifies to \\
\begin{equation}
\label{app: eq: cste}
\min_{\boldsymbol{\theta}} \{ \mathcal{L}(F(\mathbf{x},y),y) + w \cdot \| p(\bx) - \hat{z}(\bx) \|_{2}^{2}\ + \max_{\tilde{\mathbf{x}}\in \mathcal{B}_{\epsilon}(\bx)} \{\mathcal{L}(F(\tilde{\mathbf{x}},y),y) + w \cdot \| p(\tilde{\bx}) - \hat{z}(\bx) \|_{2}^{2}\}\} 
\end{equation}
which we denote as \ATCS. Similarly, \MLCATBS \ objective becomes \MLCATCS \ by including TE and clean input sample as 
\begin{equation}
\label{app: eq: cswp}
\min_{\boldsymbol{\theta}} \{  \mathcal{L}(F_{\theta + v}(\mathbf{x},y),y) + w \cdot \| p(\bx) - \hat{z}(\bx) \|_{2}^{2} \ + \ \max_{\mathbf{v} \in \mathcal{V}} \max_{\tilde{\mathbf{x}}\in \mathcal{B}_{\epsilon}(\bx)} \{\mathcal{L}(F_{\theta + v}(\tilde{\mathbf{x}},y),y) + w \cdot \| p(\tilde{\bx}) - \hat{z}(\bx) \|_{2}^{2}\}\} 
\end{equation}

We also consider the case when MLCAT-WP objective is modified to include cross entropy loss on boundary sample $\bx'$ but no temporal ensembling is employed. We denote this scheme as \MLCATXB  \ which is given by
\begin{equation}
\label{app: eq: xbwp}
\min_{\boldsymbol{\theta}} \{  \mathcal{L}(F_{\theta + v}(\mathbf{x}',y),y) \ + \ \max_{\mathbf{v} \in \mathcal{V}} \max_{\tilde{\mathbf{x}}\in \mathcal{B}_{\epsilon}(\bx)} \{\mathcal{L}(F_{\theta + v}(\tilde{\mathbf{x}},y),y)\}\} 
\end{equation}

\begin{figure}[h!]
\centering
\subfloat[Test Data Accuracy]{\includegraphics[width=8cm]{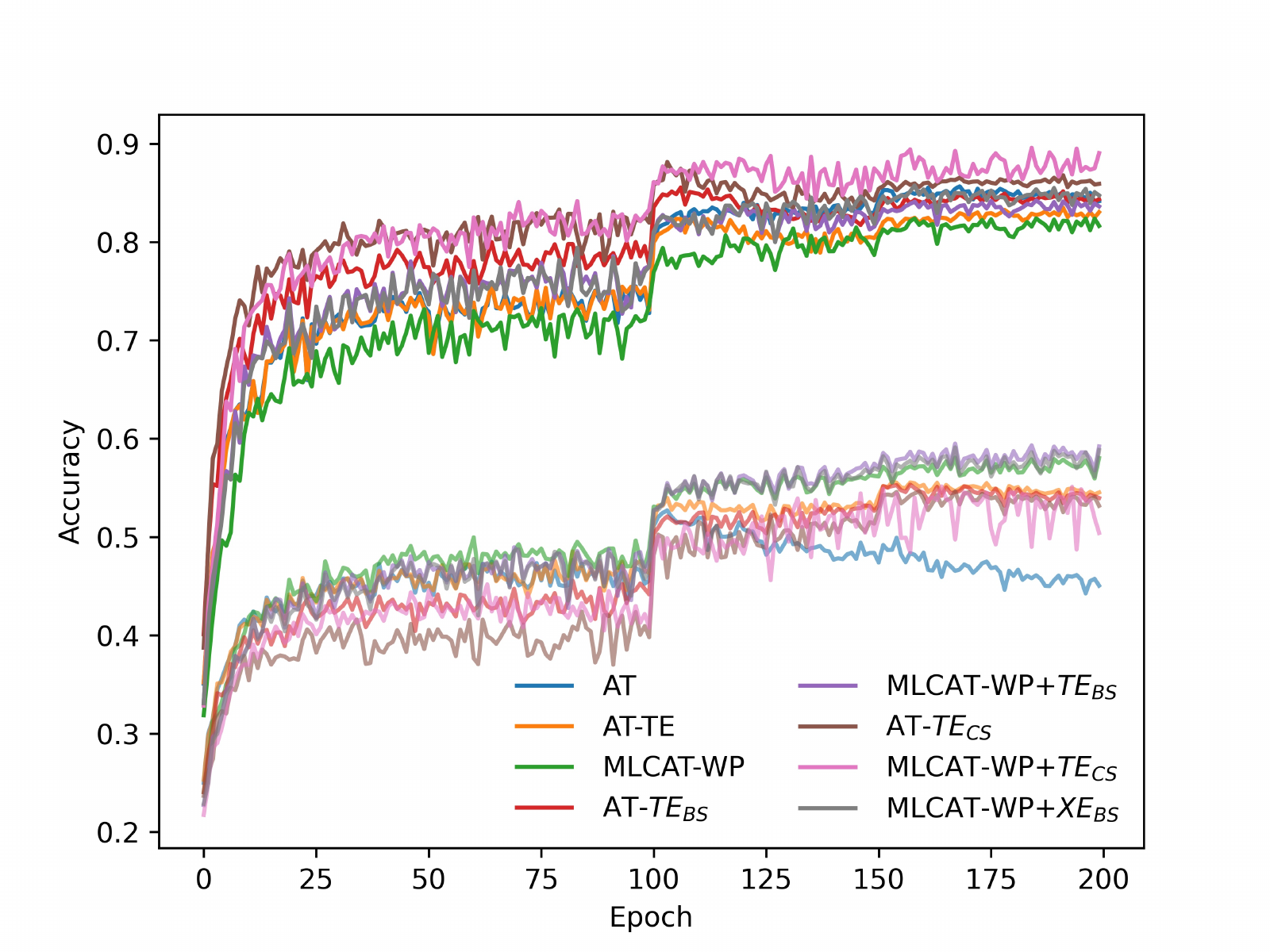}}
\subfloat[TCP]{\includegraphics[width=8cm]{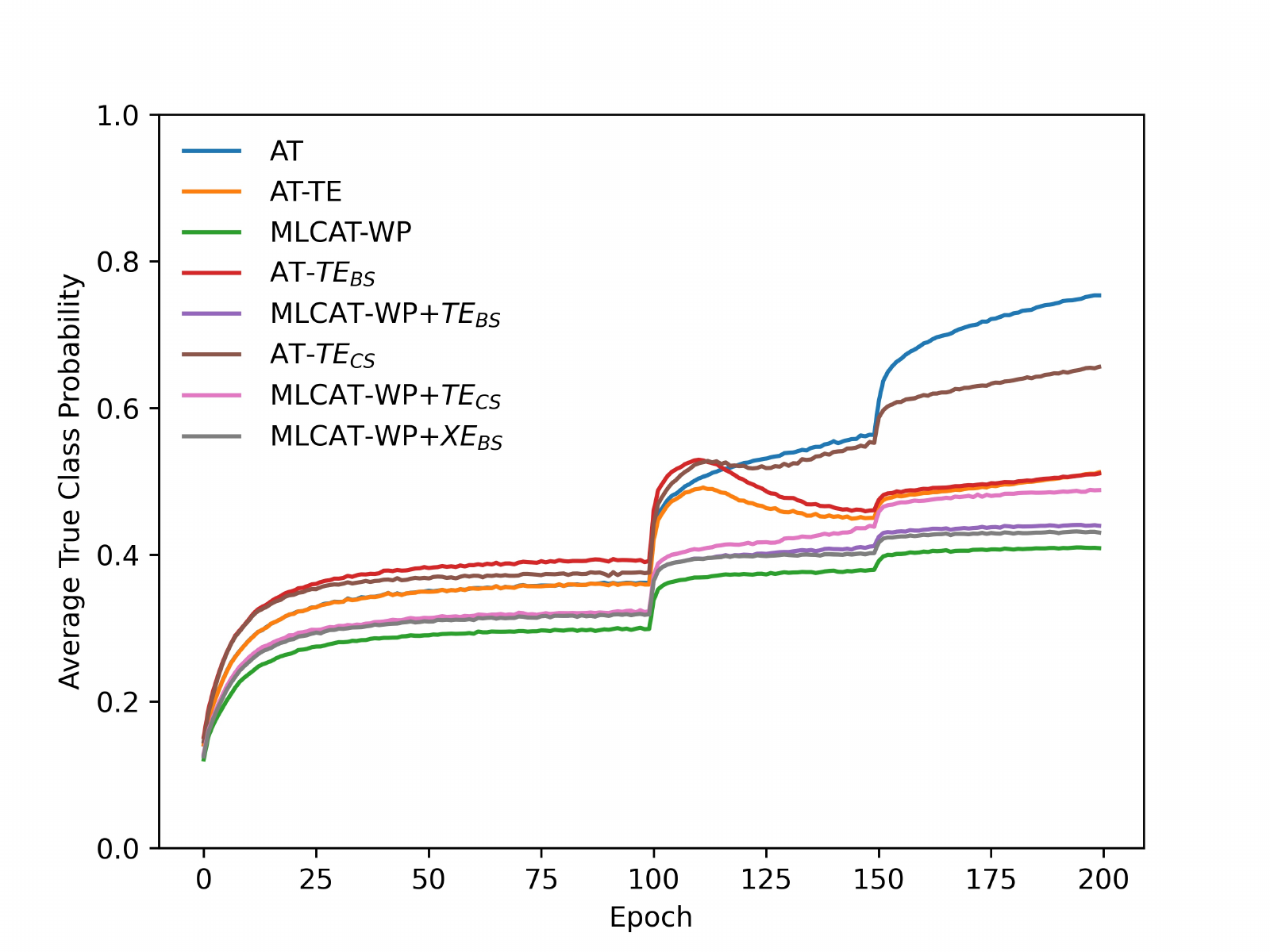}}\\
\subfloat[TCP, loss $\in$ [0, 0.5)]{\includegraphics[width=8cm]{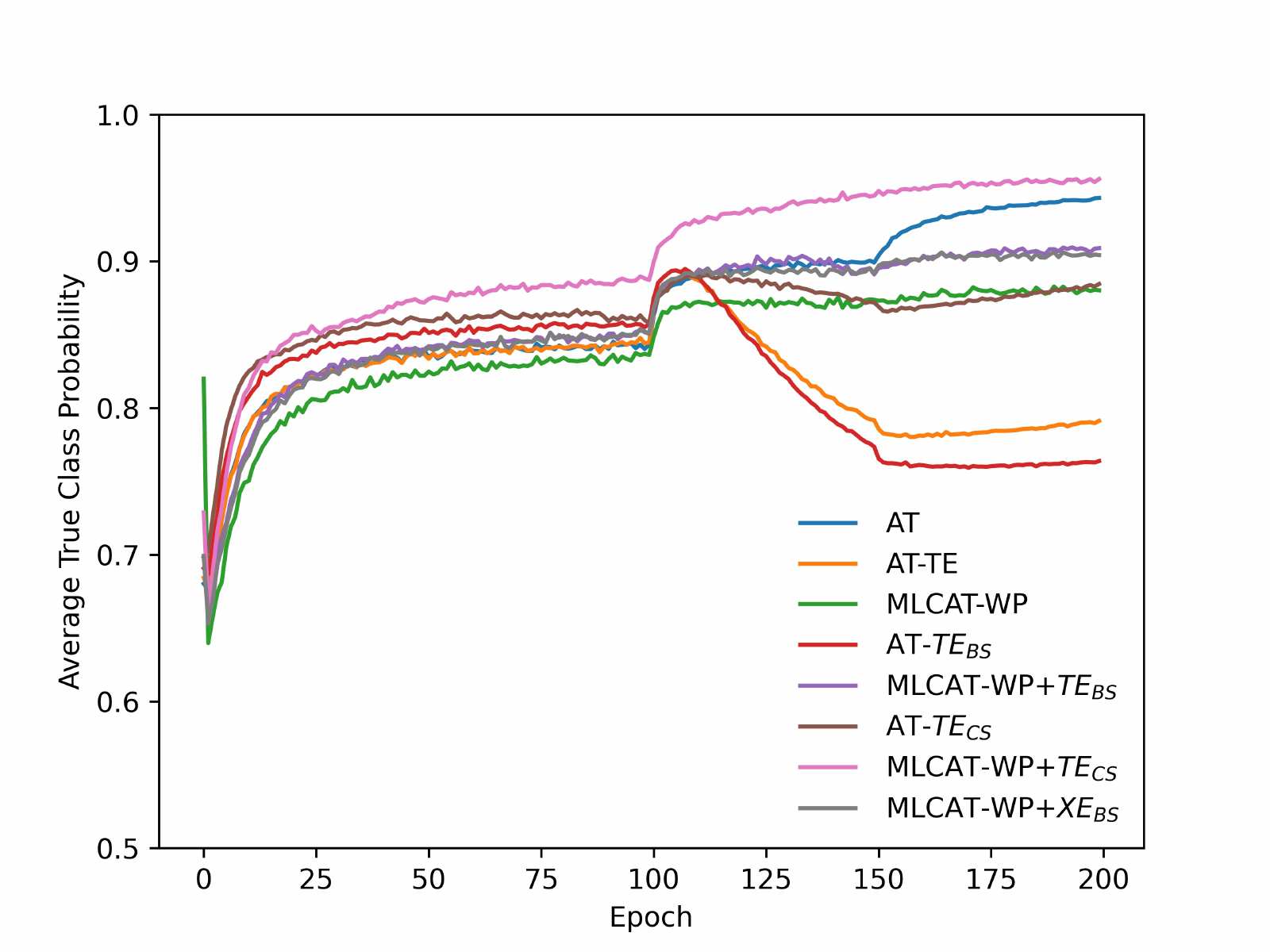}}
\subfloat[Sample Proportion, loss $\in$ [0, 0.5)]{\includegraphics[width=8cm]{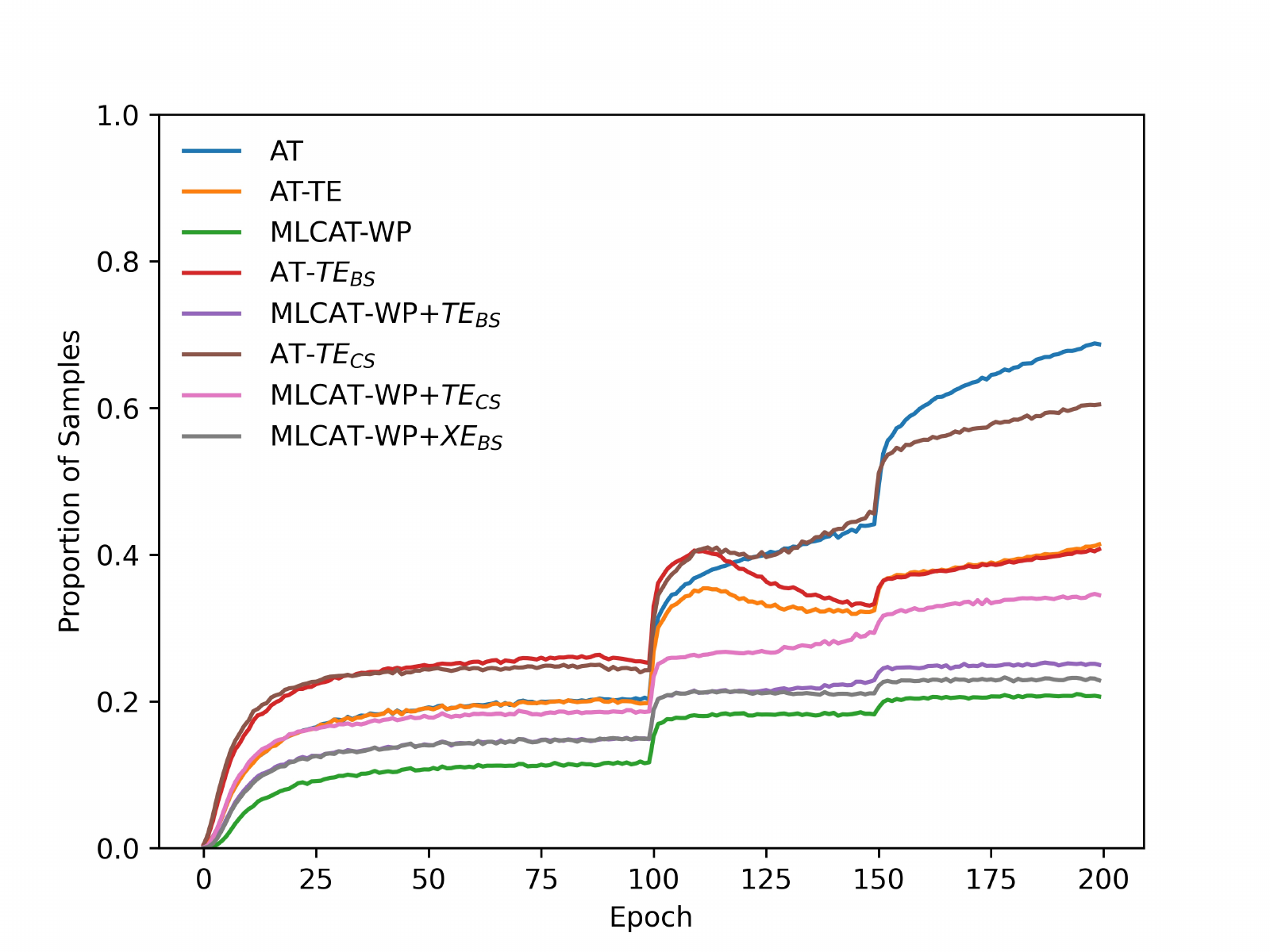}}\vspace{-1mm}

\caption{CIFAR-10 training for ResNet-18. (a) Test accuracy against clean data (dark solid lines) and $\PGDTwe$ attack (dim solid lines) are plotted.}
\label{app: fig: cifar10c}
\vspace{-6mm}
\end{figure}

\begin{figure}[h!]
\centering
\subfloat[Test Data Accuracy]{\includegraphics[width=8cm]{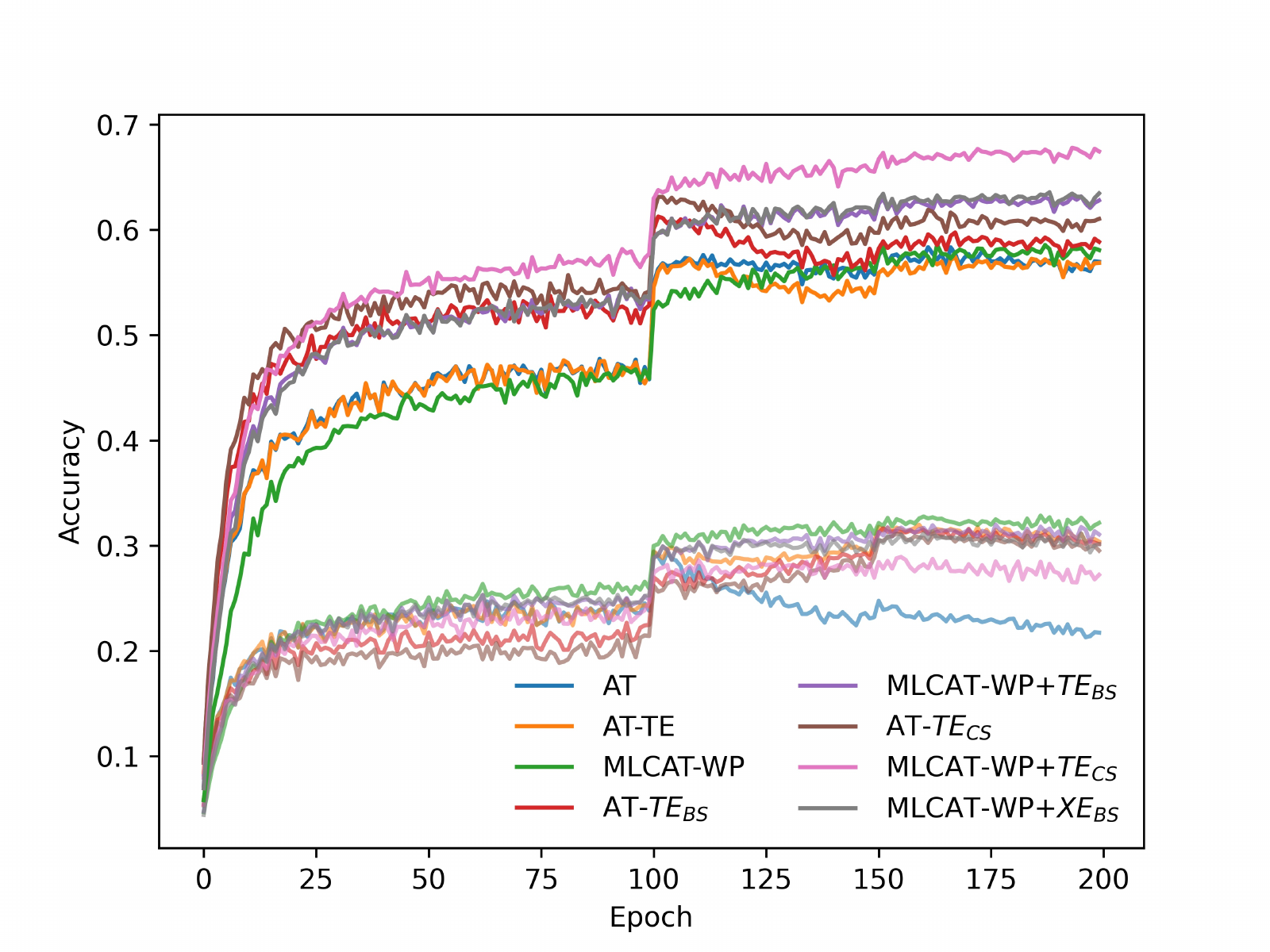}}
\subfloat[TCP]{\includegraphics[width=8cm]{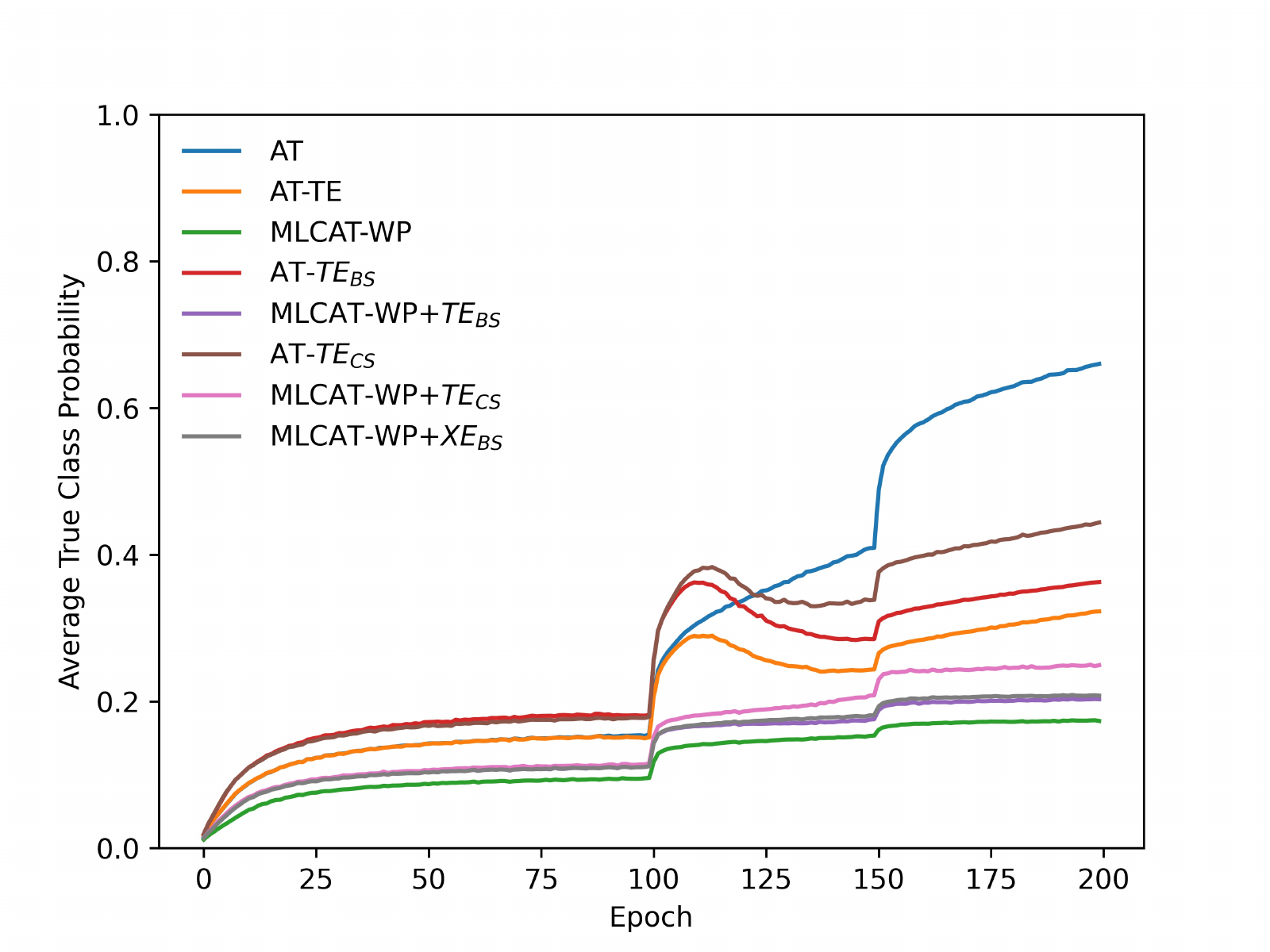}}\\
\subfloat[TCP, loss $\in$ [0, 1.0)]{\includegraphics[width=8cm]{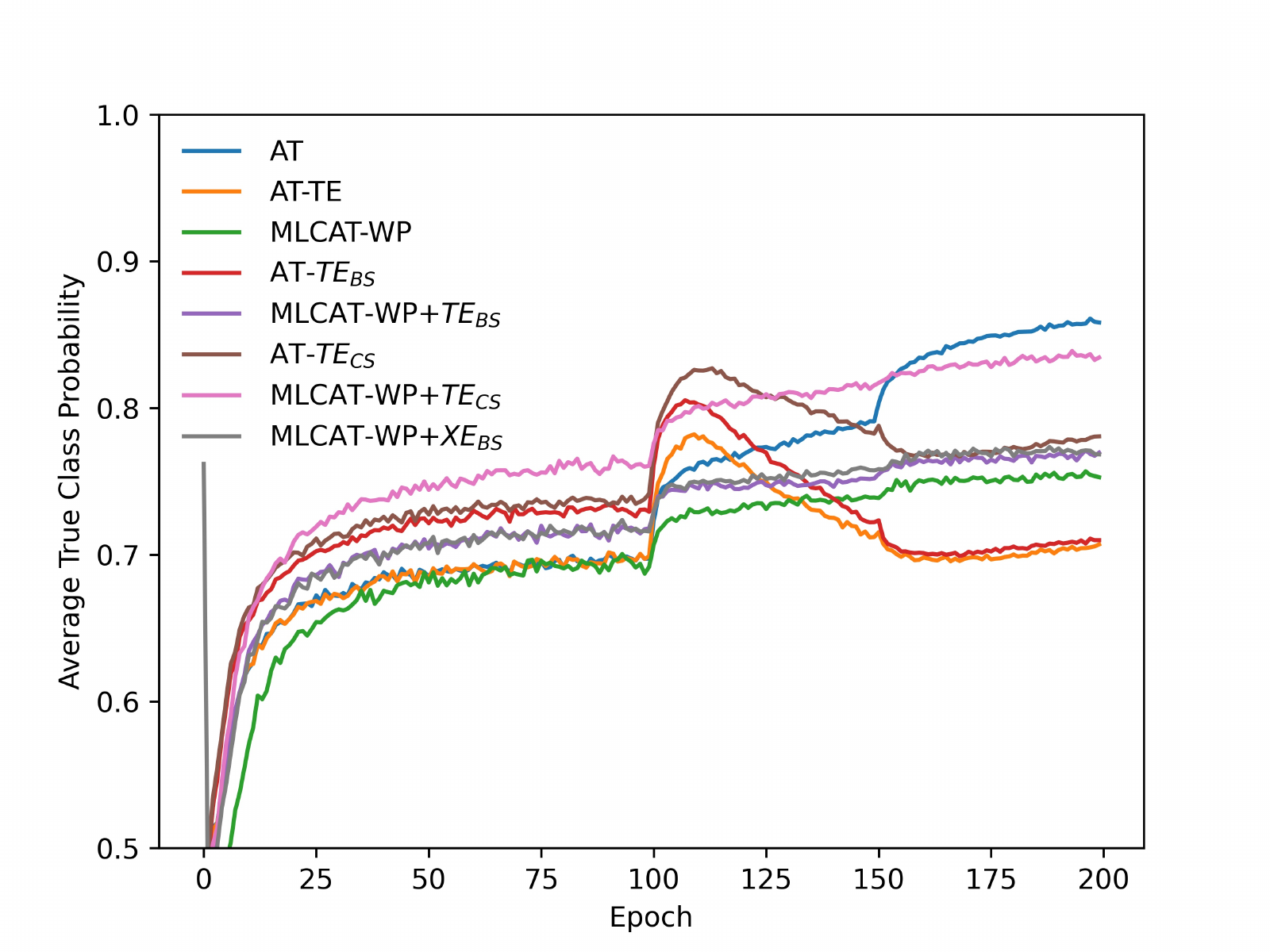}}
\subfloat[Sample Proportion, loss $\in$ [0, 1.0)]{\includegraphics[width=8cm]{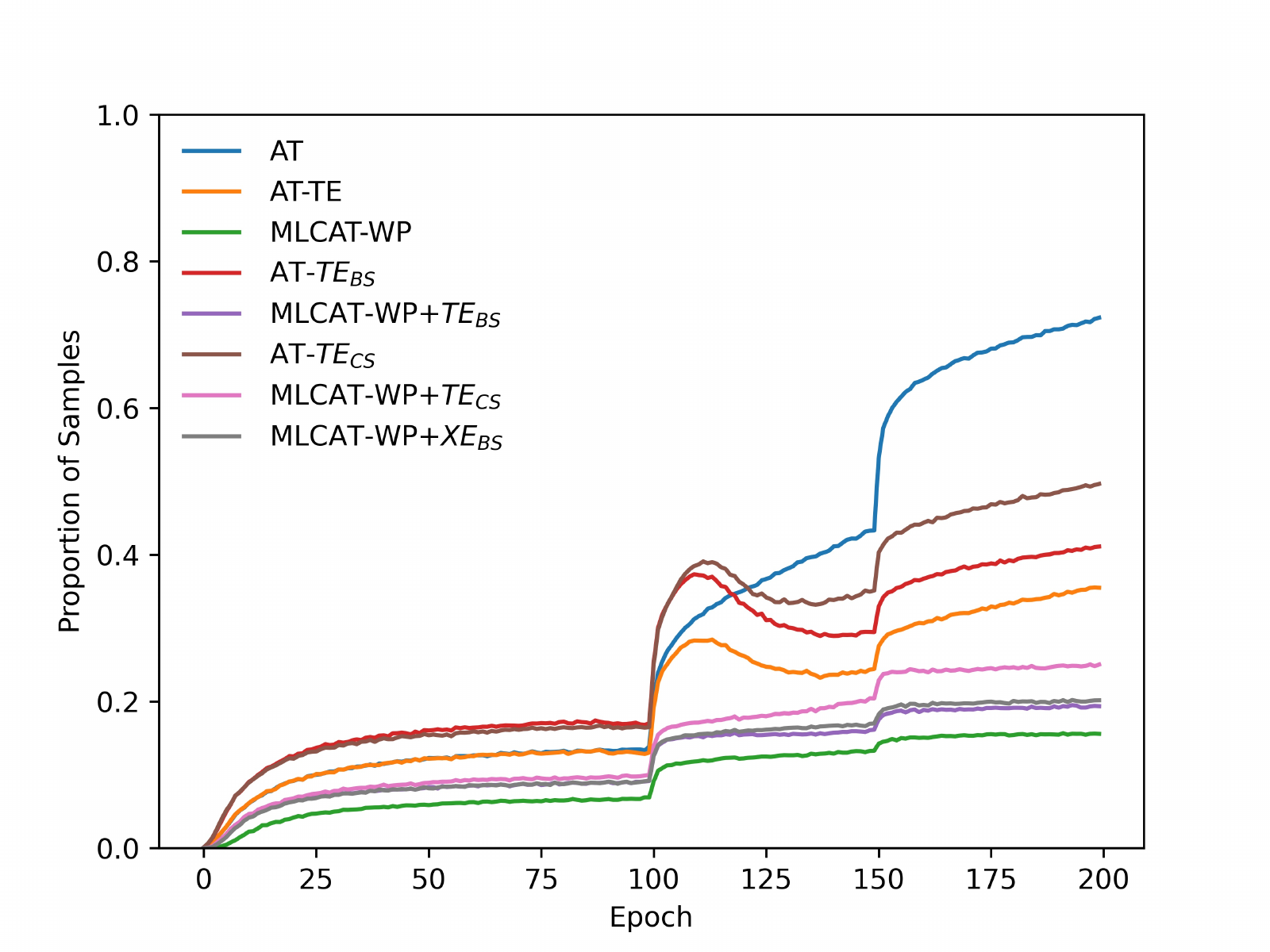}}\vspace{-1mm}

\caption{CIFAR-100 training for ResNet-18. (a) Test accuracy against clean data (dark solid lines) and $\PGDTwe$ attack (dim solid lines) are plotted.}
\label{app: fig: cifar100c}
\vspace{-6mm}
\end{figure}

\begin{figure}[h!]
\centering
\subfloat[Test Data Accuracy]{\includegraphics[width=8cm]{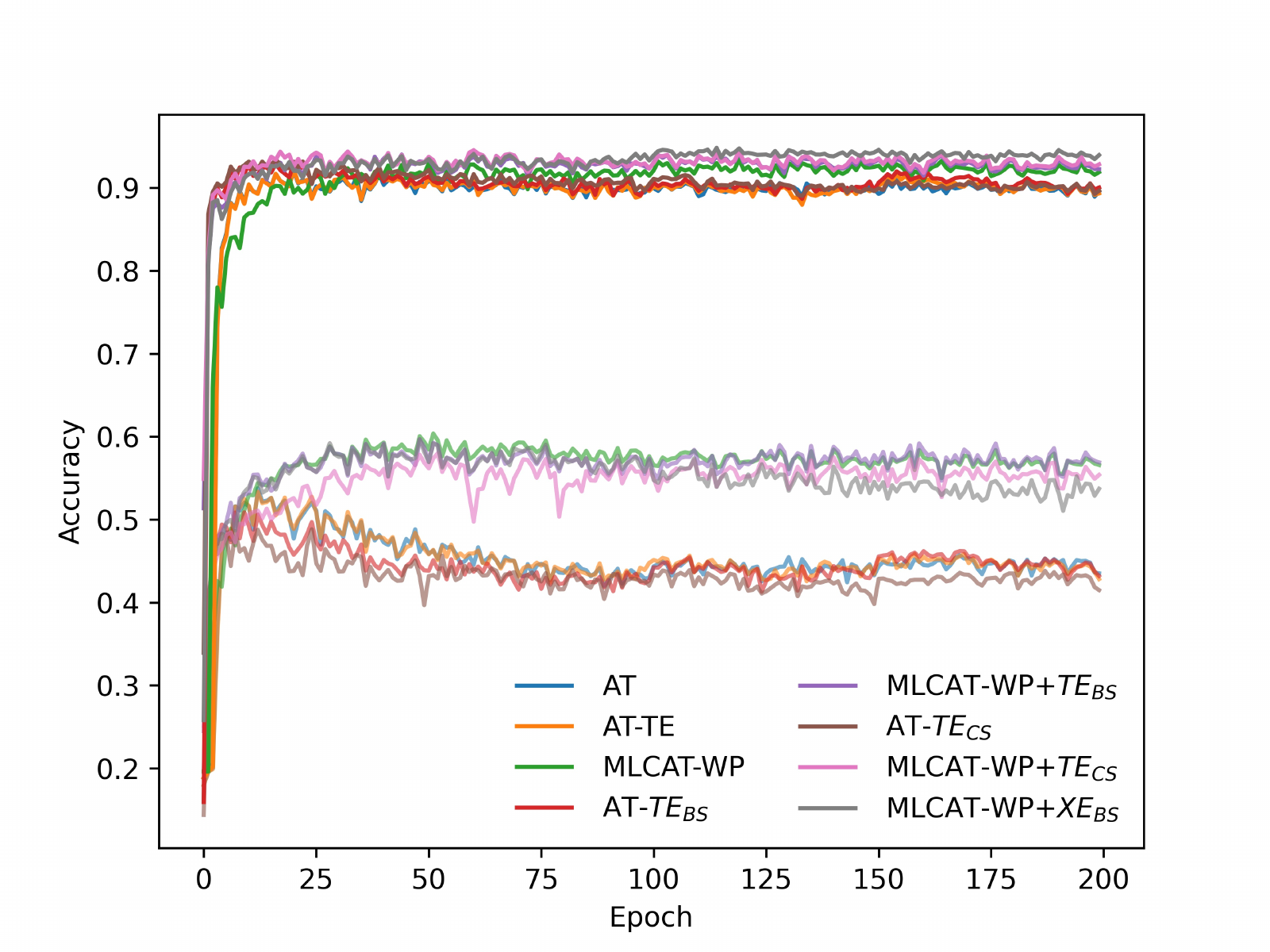}}
\subfloat[TCP]{\includegraphics[width=8cm]{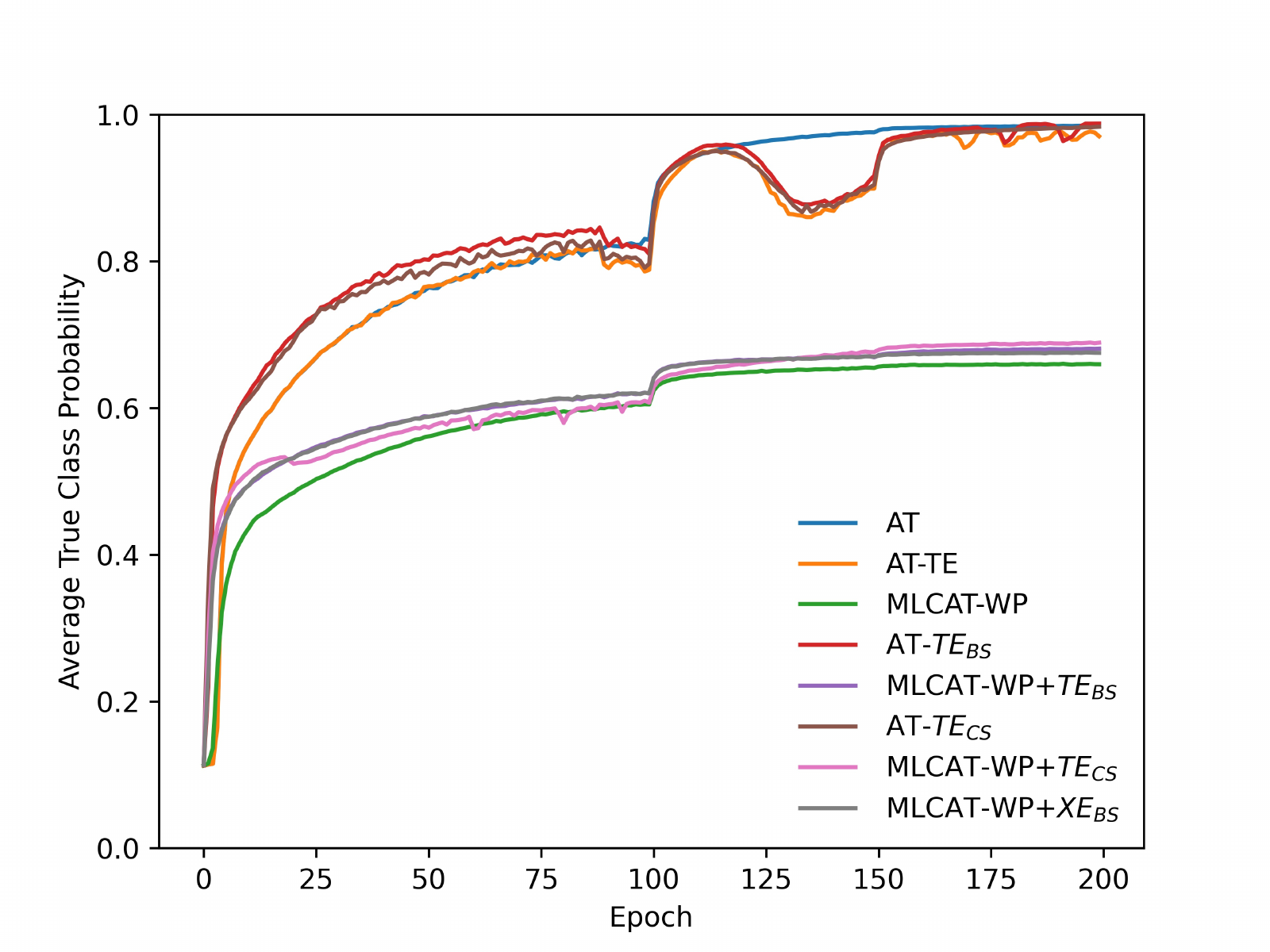}}\\
\subfloat[TCP, loss $\in$ [0, 0.5)]{\includegraphics[width=8cm]{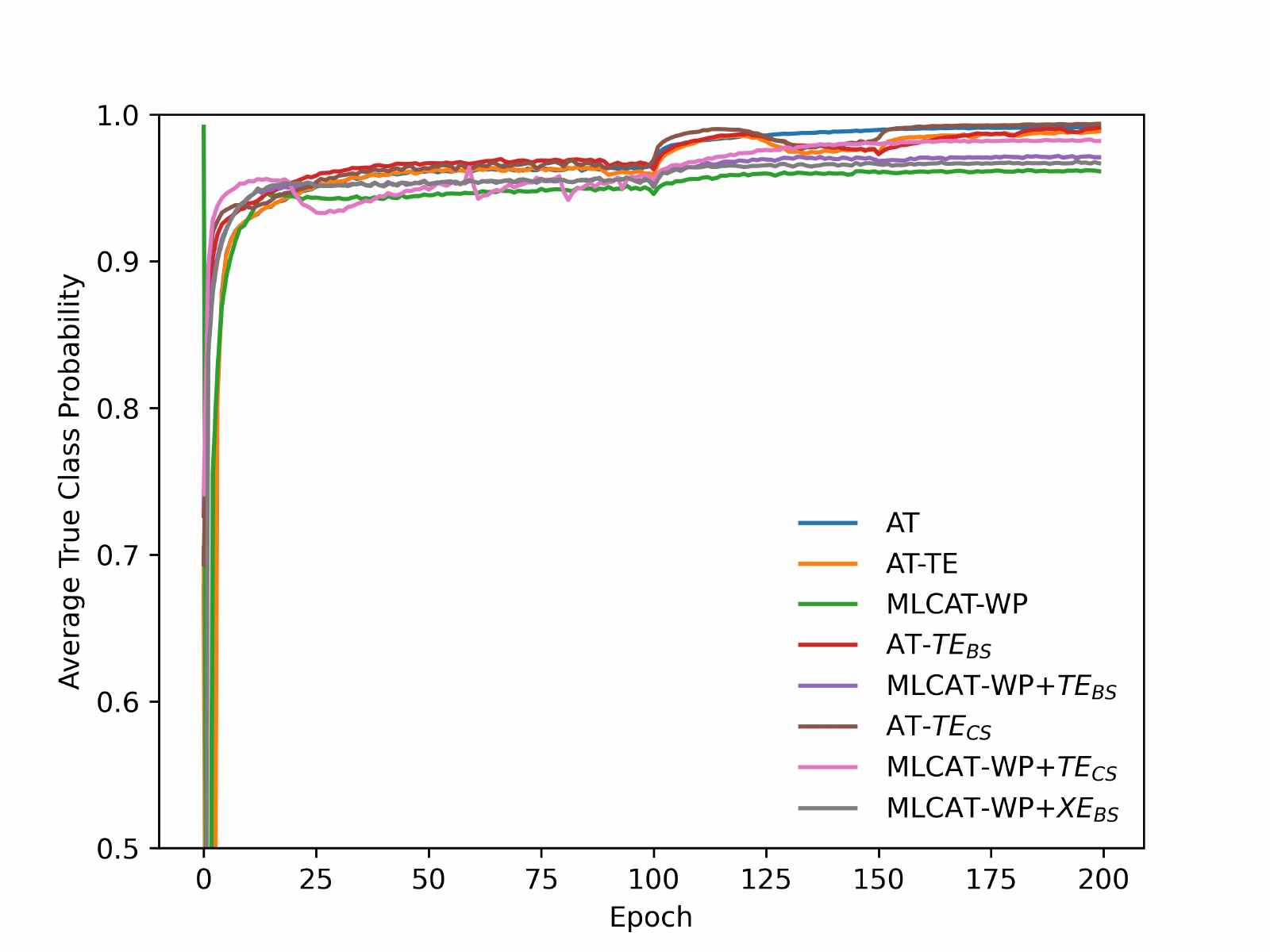}}
\subfloat[Sample Proportion, loss $\in$ [0, 0.5)]{\includegraphics[width=8cm]{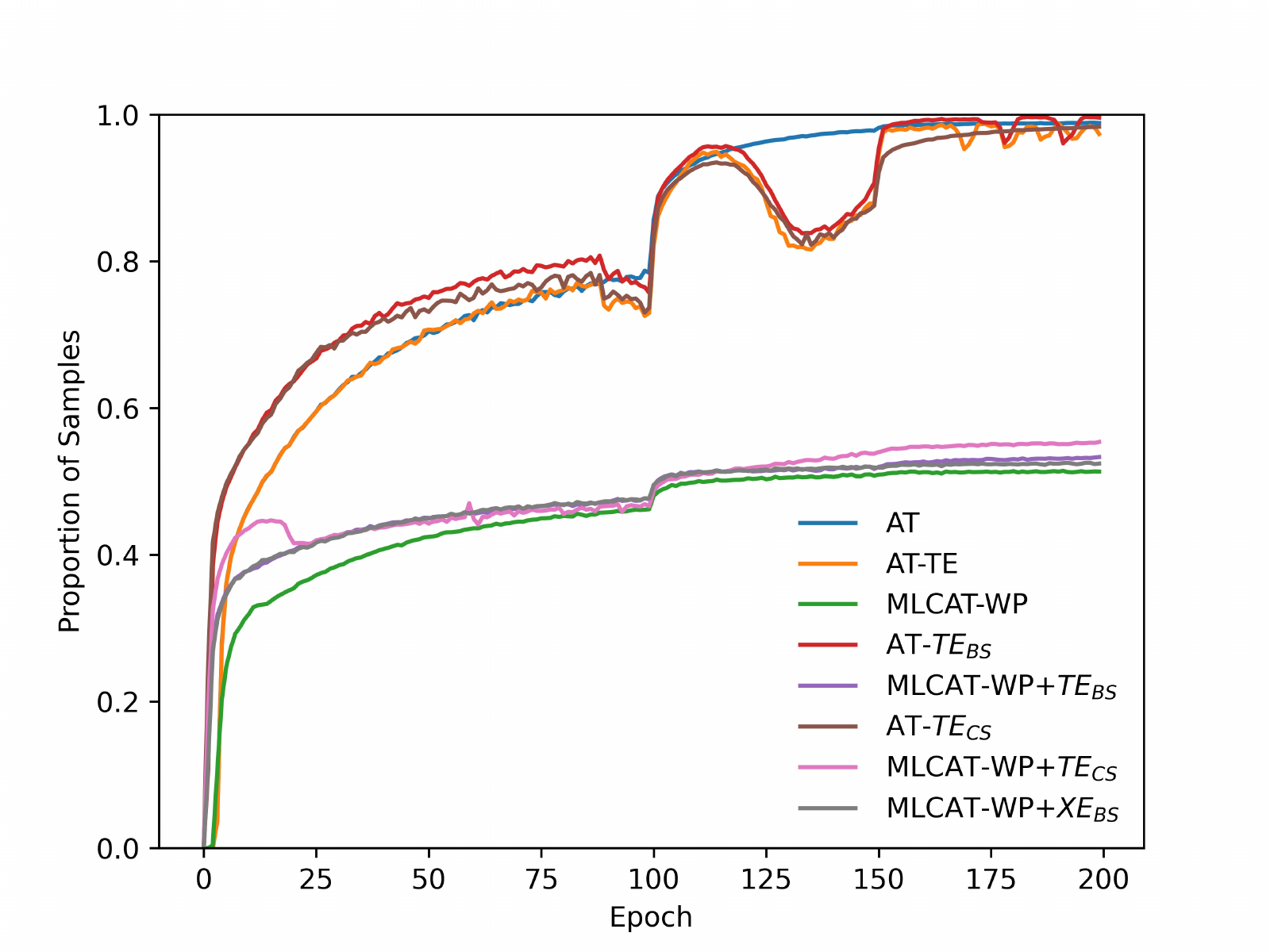}}\vspace{-1mm}

\caption{SVHN training for ResNet-18. (a) Test accuracy against clean data (dark solid lines) and $\PGDTwe$ attack (dim solid lines) are plotted.}
\label{app: fig: svhnc}
\vspace{-6mm}
\end{figure}

Table~\ref{app: Tab: tab1} shows the inherent trade off between robust accuracy and clean accuracy when clean input sample is used in place of boundary sample. Adversarial weight perturbation based approaches \MLCATBS \ and \MLCATCS \ seem to be more sensitive to the choice of sample as boundary sample clearly leads to significantly higher robust accuracy whereas clean input sample leads to significantly higher clean accuracy. In addition, robust overfitting occurs more severely for \MLCATCS \ compared to \MLCATBS. \MLCATXB \ that does not employ temporal ensembling attains clean and robust accuracy that lie in between \MLCATBS \ and \MLCATCS. On the other hand, \ATCS \ with clean input sample has increased clean accuracy and comparable robust accuracy to \ATBS \ for CIFAR-10 and CIFAR-100 datasets. Both \ATBS \ and \ATCS \ suffer from robust overfitting for SVHN dataset training and the reason is that the networks learn data in few epochs and start to assign high confidence predictions to input samples as shown in Figure~\ref{app: fig: svhnc}(b). Since regularization based on network current prediction and ensemble prediction activates at a later stage close to first learning rate decay at epoch 100, it becomes ineffective as the ensemble prediction is also very high similar to current prediction. Figure~\ref{app: fig: cifar10c}-Figure~\ref{app: fig: svhnc} show how accuracy, average TCP and proportion of small loss samples evolve with time for CIFAR-10, CIFAR-100 and SVHN datasets, respectively.

\begin{table}[h!]
  \caption{Test accuracy (mean and standard deviation of 5 runs). ``Last" and ``Best" refer to test accuracy at the end of training, and end of epoch that gives the highest accuracy w.r.t. test data respectively.}
  \label{app: Tab: tab1}
  \centering
\resizebox{0.8\columnwidth}{!}{%
  \begin{tabular}{llllllll}
    \toprule
    Dataset &   Name    &   Clean(Best) &   Clean(Last)    &  Robust(Best)  &   Robust(Last)    &   AA(Best)    &   AA(Last) \\
    \midrule
    \multirow{5}*{CIFAR-10} & \ATBS     &   83.99$\pm$0.14 & 84.46$\pm$0.36 &   55.26$\pm$0.07    &   53.58$\pm$0.74    &  50.17$\pm$0.09   &  48.80$\pm$0.70 \\
    & \ATCS     &   85.62$\pm$0.23 & 85.76$\pm$0.18 &   54.83$\pm$0.23    &   53.21$\pm$0.91    &  50.09$\pm$0.21   &  48.81$\pm$0.61 \\
    & \MLCATBS  & 83.712$\pm$0.51 &   83.75$\pm$0.41    &  \textbf{59.20$\pm$0.21}    &       \textbf{58.53$\pm$0.52}    &  \textbf{50.38$\pm$0.07}   &  \textbf{50.30$\pm$0.26}   \\
    & \MLCATCS  & \textbf{86.17$\pm$0.37} &   \textbf{88.47$\pm$0.67}    &  55.37$\pm$0.28    &       51.68$\pm$1.18    &  48.13$\pm$0.40   &  46.84$\pm$0.23   \\
    & \MLCATXB  & 84.81$\pm$0.3 &   84.91$\pm$0.33    &  58.76$\pm$0.16    &      57.8$\pm$0.61    &  50.09$\pm$0.09   &  49.75$\pm$0.32   \\
    \midrule    
    \multirow{5}*{CIFAR-100} & \ATBS     &   58.78$\pm$0.23  &   58.82$\pm$0.29  &   31.59$\pm$0.09  &   30.40$\pm$0.16  &  25.99$\pm$0.18  &   24.98$\pm$0.23      \\
    & \ATCS     &   60.71$\pm$0.37  &   60.52$\pm$0.41  &   31.08$\pm$0.22  &   29.72$\pm$0.19  &  25.72$\pm$0.23  &   24.77$\pm$0.15      \\
    & \MLCATBS     &    62.50$\pm$0.28  &   62.45$\pm$0.42  &   \textbf{31.95$\pm$0.06}  &   \textbf{30.72$\pm$0.65}  &   \textbf{26.57$\pm$0.14} &  \textbf{25.74$\pm$0.48}  \\
    & \MLCATCS     &    \textbf{66.38$\pm$0.82}  &   \textbf{67.25$\pm$0.38}  &   28.87$\pm$0.25  &   26.96$\pm$0.53  &   24.00$\pm${0.26} &  22.59$\pm$0.62  \\
    & \MLCATXB     &    62.67$\pm$0.3  &   63.1$\pm$0.37  &   31.19$\pm$0.15  &   29.95$\pm$0.51  &   26.29$\pm$0.14 &  25.46$\pm$0.49  \\
    \midrule    
    \multirow{5}*{SVHN}    & \ATBS     & 91.58$\pm$0.32  &   90.43$\pm$0.46  &   50.94$\pm$0.16  &   44.43$\pm$0.66    &  45.37$\pm$0.29  &   39.43$\pm$0.70      \\
    & \ATCS     & 91.75$\pm$0.69  &   90.24$\pm$0.43  &   48.72$\pm$0.45  &   43.02$\pm$0.79    &  42.42$\pm$0.27  &   38.22$\pm$0.68      \\
    & \MLCATBS     &    92.43$\pm$0.46  &   92.53$\pm$0.35  &   \textbf{60.07$\pm$0.31}  &   \textbf{57.61$\pm$ 0.94}    &   51.27$\pm$0.49  &   \textbf{49.42$\pm$0.54}  \\
    & \MLCATCS     &    \textbf{93.18$\pm$0.69}  &   \textbf{93.57$\pm$0.41}  &   56.22$\pm$1.17  &   52.05$\pm$ 1.78    &   48.71$\pm$1.02  &   45.19$\pm$1.01  \\
    & \MLCATXB     &    92.47$\pm$0.44  &   93.4$\pm$0.82  &   59.78$\pm$0.04  &   55.21$\pm$ 1.2    &   \textbf{51.29$\pm$0.24}  &   46.73$\pm$1.81  \\
    \bottomrule
  \end{tabular}}
\vspace{-3mm}
\end{table}

\end{document}